# Graph Kernels

**S. V. N. Vishwanathan**                                    SVN.Vishwanathan@nicta.com.au
*College of Engineering and Computer Science*
*Australian National University* and *NICTA*
*Locked Bag 8001, Canberra ACT 2601, Australia*

**Karsten M. Borgwardt**                                            kmb51@cam.ac.uk
*Machine Learning Group, Department of Engineering*
*University of Cambridge*
*Trumpington Street, CB2 1PZ Cambridge, United Kingdom*

**Imre Risi Kondor**                                        risi@gatsby.ucl.ac.uk
*Gatsby Computational Neuroscience Unit*
*University College London*
*17 Queen Square, WC1N 3AR London, United Kingdom*

**Nicol N. Schraudolph**                                        jmlr@schraudolph.org
*College of Engineering and Computer Science*
*Australian National University* and *NICTA*
*Locked Bag 8001, Canberra ACT 2601, Australia*

## Abstract

We present a unified framework to study graph kernels, special cases of which include the random walk graph kernel (Gärtner et al., 2003; Borgwardt et al., 2005), marginalized graph kernel (Kashima et al., 2003, 2004; Mahé et al., 2004), and geometric kernel on graphs (Gärtner, 2002). Through extensions of linear algebra to Reproducing Kernel Hilbert Spaces (RKHS) and reduction to a Sylvester equation, we construct an algorithm that improves the time complexity of kernel computation from $O(n^6)$ to $O(n^3)$. When the graphs are sparse, conjugate gradient solvers or fixed-point iterations bring our algorithm into the sub-cubic domain. Experiments on graphs from bioinformatics and other application domains show that it is often more than a thousand times faster than previous approaches. We then explore connections between diffusion kernels (Kondor and Lafferty, 2002), regularization on graphs (Smola and Kondor, 2003), and graph kernels, and use these connections to propose new graph kernels. Finally, we show that rational kernels (Cortes et al., 2002, 2003, 2004) when specialized to graphs reduce to the random walk graph kernel.

**Keywords:** Graph kernels, Linear Algebra in RKHS, Sylvester Equations, Bioinformatics, Rational Kernels, Transducers, Semirings, Diffusion, Random Walks, Regularization on Graphs.

## 1. Introduction

We begin by providing some background, establishing some basic notation, and giving an outline of the paper.

### 1.1 Background

Machine learning in domains such as bioinformatics (Sharan and Ideker, 2006), chemoinformatics (Bonchev and Rouvray, 1991), drug discovery (Kubinyi, 2003), web data mining (Washio and Motoda, 2003), and social networks (Kumar et al., 2006) involves the study of relationships between structured objects. Graphs are natural data structures to model such structures, with nodes representing objects and edges the relations between them. In this context, one often encounters two questions: "How similar are two nodes in a given graph?" and "How similar are two graphs to each other?"

Kernel methods (Schölkopf and Smola, 2002) offer a natural framework to study these questions. Roughly speaking, a kernel $k(x, x')$ is a measure of similarity between objects $x$ and $x'$. It must satisfy two mathematical requirements: it must be symmetric, that is, $k(x, x') = k(x', x)$, and positive semi-definite (*p.s.d.*). Comparing nodes in a graph involves constructing a kernel between nodes, while comparing graphs involves constructing a kernel between graphs. In both cases, the challenge is to define a kernel that captures the semantics inherent in the graph structure but at the same time is reasonably efficient to evaluate.

Until now, the above two types of kernels have largely been studied separately. The idea of constructing kernels on graphs (*i.e.,* between the nodes of a single graph) was first proposed by Kondor and Lafferty (2002), and extended by Smola and Kondor (2003). Kernels between graphs were proposed by Gärtner (2002) (geometric kernels on graphs) and Gärtner et al. (2003) (random walk graph kernels), and later extended by Borgwardt et al. (2005). Much at the same time, the idea of marginalized kernels (Tsuda et al., 2002) was extended to graphs by Kashima et al. (2003, 2004), and further refined by Mahé et al. (2004). A seemingly independent line of research investigates the so-called rational kernels, which are kernels between finite state automata based on the algebra of abstract semirings (Cortes et al., 2004, 2003, 2002).

The aim of this paper is twofold: on one hand we present theoretical results showing that these four strands of research are in fact closely related, on the other we present new algorithms for efficiently computing kernels between graphs. Towards this end we first establish some notation and review pertinent concepts from linear algebra and graph theory.

### 1.2 Linear Algebra Concepts

We use $\mathbf{e}_i$ to denote the $i^{\text{th}}$ standard basis vector (that is, a vector of all zeros with the $i^{\text{th}}$ entry set to one), $\mathbf{e}$ to denote a vector with all entries set to one, $\mathbf{0}$ to denote the vector of all zeros, and $\mathbf{I}$ to denote the identity matrix. When it is clear from context we will not mention the dimensions of these vectors and matrices.

**Definition 1** *Given real matrices $A \in \mathbb{R}^{n \times m}$ and $B \in \mathbb{R}^{p \times q}$, the Kronecker product $A \otimes B \in \mathbb{R}^{np \times mq}$ and column-stacking operator $\mathrm{vec}(A) \in \mathbb{R}^{nm}$ are defined as*

$$
A \otimes B := \begin{bmatrix} A_{11}B & A_{12}B & \dots & A_{1m}B \\ \vdots & \vdots & \vdots & \vdots \\ A_{n1}B & A_{n2}B & \dots & A_{nm}B \end{bmatrix}, \quad \mathrm{vec}(A) := \begin{bmatrix} A_{*1} \\ \vdots \\ A_{*m} \end{bmatrix},
$$

*where $A_{*j}$ denotes the $j^{\mathrm{th}}$ column of $A$.*

Kronecker product and vec operator are linked by the well-known property (*e.g.,* Bernstein, 2005, proposition 7.1.9):

$$
\mathrm{vec}(ABC) = (C^\top \otimes A)\,\mathrm{vec}(B). \tag{1}
$$

Another well-known property of the Kronecker product, which we use in Section 5, is (Bernstein, 2005, proposition 7.1.6):

$$
(A \otimes B)(C \otimes D) = AC \otimes BD. \tag{2}
$$

A closely related concept is that of the Kronecker sum which is defined for real matrices $A \in \mathbb{R}^{n \times m}$ and $B \in \mathbb{R}^{p \times q}$ as

$$
A \oplus B := A \otimes \mathbf{I}_{pq} + \mathbf{I}_{nm} \otimes B, \tag{3}
$$

with $\mathbf{I}_{nm}$ (*resp.* $\mathbf{I}_{pq}$) denoting the $n \times m$ (*resp.* $p \times q$) identity matrix. Many of its properties can be derived from those of the Kronecker product.

Finally, the Hadamard product of two real matrices $A, B \in \mathbb{R}^{n \times m}$, denoted by $A \odot B \in \mathbb{R}^{n \times m}$, is obtained by element-wise multiplication. It interacts with the Kronecker product via

$$
(A \otimes B) \odot (C \otimes D) = (A \odot C) \otimes (B \odot D). \tag{4}
$$

All the above concepts can be extended to a Reproducing Kernel Hilbert Space (RKHS) (See Appendix A for details).

## 1.3 Graph Concepts

A graph $G$ consists of an ordered set of $n$ vertices $V = \{v_1, v_2, \dots, v_n\}$, and a set of edges $E \subset V \times V$. A vertex $v_i$ is said to be a neighbor of another vertex $v_j$ if they are connected by an edge, *i.e.,* if $(v_i, v_j) \in E$; this is also denoted $v_i \sim v_j$. A walk of length $t$ on $G$ is a sequence of indices $i_1, i_2, \dots i_{t+1}$ such that $v_{i_k} \sim v_{i_{k+1}}$ for all $1 \leq k \leq t$. A graph is said to be connected if any two pairs of vertices can be connected by a walk. In this paper we will always work with connected graphs. A graph is said to be undirected if $(v_i, v_j) \in E \iff (v_j, v_i) \in E$.

In much of the following we will be dealing with weighted graphs, which are a slight generalization of the above. In a weighted graph, each edge $(v_i, v_j)$ has an associated weight $w_{ij} > 0$ signifying its "strength". If $v_i$ and $v_j$ are not neighbors, then $w_{ij} = 0$. In an undirected weighted graph $w_{ij} = w_{ji}$.

The adjacency matrix of an unweighted graph is an $n \times n$ matrix $\widetilde{\mathrm{A}}$ with $\widetilde{\mathrm{A}}_{ij} = 1$ if $v_i \sim v_j$, and 0 otherwise. The adjacency matrix of a weighted graph is just the matrix of weights, $\widetilde{\mathrm{A}}_{ij} = w_{ij}$. In both cases, if $G$ is undirected, then the adjacency matrix is symmetric. The diagonal entries of $\widetilde{\mathrm{A}}$ are always zero.

The adjacency matrix has a normalized cousin, defined $A := D^{-1} \widetilde{\mathrm{A}}$, which has the property that each of its rows sums to one, therefore it can serve as the transition matrix for a stochastic process. Here, $D$ is a diagonal matrix of node degrees, $d_i$, such that $D_{ii} = d_i = \sum_j \widetilde{\mathrm{A}}_{ij}$. A random walk on $G$ is a process generating sequences of vertices $v_{i_1}, v_{i_2}, v_{i_3}, \ldots$ according to $\mathbb{P}(i_{k+1}|i_1, \ldots i_k) = A_{i_k, i_{k+1}}$, that is, the probability at $v_{i_k}$ of picking $v_{i_{k+1}}$ next is proportional to the weight of the edge $(v_{i_k}, v_{i_{k+1}})$. The $t^{\mathrm{th}}$ power of $A$ thus describes $t$-length walks, $i.e.,$ $[A^t]_{ij}$ is the probability of a transition from vertex $v_i$ to vertex $v_j$ via a walk of length $t$. If $p_0$ is an initial probability distribution over vertices, the probability distribution $p_t$ describing the location of our random walker at time $t$ is $p_t = A^t p_0$. The $j^{\mathrm{th}}$ component of $p_t$ denotes the probability of finishing a $t$-length walk at vertex $v_j$. We will use this intuition to define generalized random walk graph kernels.

Let $\mathcal{X}$ be a set of labels which includes the special label $\zeta$. Every edge-labeled graph $G$ is associated with a label matrix $X \in \mathcal{X}^{n \times n}$ such that $X_{ij} = \zeta$ iff $(v_i, v_j) \notin E$, in other words only those edges which are present in the graph get a non-$\zeta$ label. Let $\mathcal{H}$ be the RKHS endowed with the kernel $\kappa : \mathcal{X} \times \mathcal{X} \to \mathbb{R}$, and let $\phi : \mathcal{X} \to \mathcal{H}$ denote the corresponding feature map which maps $\zeta$ to the zero element of $\mathcal{H}$. We use $\Phi(X)$ to denote the feature matrix of $G$ (see Appendix A for details). For ease of exposition we do not consider labels on vertices here, though our results hold for that case as well. Henceforth we use the term labeled graph to denote an edge-labeled graph.

## 1.4 Paper Outline

In the first part of this paper (Sections 2–4) we present a unifying framework for graph kernels, encompassing many known kernels as special cases, and connecting to others. We describe our framework in Section 2, prove that it leads to $p.s.d.$ kernels, and discuss random walk graph kernels, geometric kernels on graphs, and marginalized graph kernels as special cases. For ease of exposition we will work with real matrices in the main body of the paper and relegate the RKHS extensions to Appendix A. In Section 3 we present three efficient ways to compute random walk graph kernels, namely 1. via reduction to a Sylvester equation, 2. using a conjugate gradient (CG) solver, and 3. using a fixed point iteration. Experiments on a variety of real and synthetic datasets in Section 4 illustrate the computational advantages of our approach, which reduces the time complexity of kernel computations from $O(n^6)$ to $O(n^3)$.

In the second part (Sections 5–7) we draw further connections to existing kernels on structured objects. In Section 5 we present a simple proof that rational kernels are $p.s.d.$, and show that specializing them to graphs yields random walk graph kernels. In Section 6 we discuss the relation between R-convolution kernels and various incarnations of graph kernels. In fact, all known graph kernels can be shown to be instances of R-convolution kernels. We also show that extending the framework therough the use of semirings does not always result in a $p.s.d.$ kernel; a case in point is the optimal assignment kernel of Fröhlich et al. (2006). In Section 7 we shift our attention to diffusion processes on graphs and associated kernels;

this leads us to propose new kernels on graphs, based on the Cartesian graph product. We show that the efficient computation techniques we introduced in Section 3 are also applicable here, but are ultimately forced to conclude that diffusion-based graph kernels are not useful in a general context. We conclude in Section 8 with an outlook and discussion.

## 2. Random Walk Graph Kernels

Our generalized random walk graph kernels are based on a simple idea: given a pair of graphs, perform random walks on both, and count the number of matching walks. We show that this simple concept underlies random walk graph kernels, marginalized graph kernels, and geometric kernels on graphs. In order to do this, we first need to introduce direct product graphs.

### 2.1 Direct Product Graphs

Given two graphs $G(V, E)$ and $G'(V', E')$ (with $|V| = n$ and $|V'| = n'$), their direct product $G_\times$ is a graph with vertex set

$$V_\times = \{(v_i, v_{i'}') : v_i \in V, \ v_{i'}' \in V'\}, \tag{5}$$

and edge set

$$E_\times = \{((v_i, v_{i'}'), (v_j, v_{j'}')) : (v_i, v_j) \in E \wedge (v_{i'}', v_{j'}') \in E'\}. \tag{6}$$

In other words, $G_\times$ is a graph over pairs of vertices from $G$ and $G'$, and two vertices in $G_\times$ are neighbors if and only if the corresponding vertices in $G$ and $G'$ are both neighbors (see Figure 1 for an illustration). If $\widetilde{A}$ and $\widetilde{A}'$ are the respective adjacency matrices of $G$ and $G'$, then the adjacency matrix of $G_\times$ is $\widetilde{A}_\times = \widetilde{A} \otimes \widetilde{A}'$. Similarly, $A_\times = A \otimes A'$.

Performing a random walk on the direct product graph is equivalent to performing a simultaneous random walk on $G$ and $G'$ (Imrich and Klavžar, 2000). If $p$ and $p'$ denote initial probability distributions over the vertices of $G$ and $G'$, then the corresponding initial probability distribution on the direct product graph is $p_\times := p \otimes p'$. Likewise, if $q$ and $q'$ are stopping probabilities (that is, the probability that a random walk ends at a given vertex), then the stopping probability on the direct product graph is $q_\times := q \otimes q'$.

If $G$ and $G'$ are edge-labeled, we can associate a weight matrix $W_\times \in \mathbb{R}^{nn' \times nn'}$ with $G_\times$ using our Kronecker product in RKHS (Definition 12): $W_\times = \Phi(X) \otimes \Phi(X')$. As a consequence of the definition of $\Phi(X)$ and $\Phi(X')$, the entries of $W_\times$ are non-zero only if the corresponding edge exists in the direct product graph. The weight matrix is closely related to the normalized adjacency matrix: assume that $\mathcal{H} = \mathbb{R}$ endowed with the usual inner product, and $\phi(X_{ij}) = 1/d_i$ if $(v_i, v_j) \in E$ or zero otherwise. Then $\Phi(X) = A$ and $\Phi(X') = A'$, and consequently $W_\times = A_\times$, that is, the weight matrix is identical to the normalized adjacency matrix of the direct product graph.

To extend the above discussion, assume that $\mathcal{H} = \mathbb{R}^d$ endowed with the usual inner product, and that there are $d$ distinct edge labels $\{1, 2, \ldots, d\}$. For each edge $(v_i, v_j) \in E$ we have $\phi(X_{ij}) = \mathbf{e}_l/d_i$ if the edge $(v_i, v_j)$ is labeled $l$. All other entries of $\Phi(X)$ are set to $\mathbf{0}$. $\kappa$ is therefore a delta kernel, that is, its value between any two edges is one iff the labels on the edges match, and zero otherwise. The weight matrix $W_\times$ has a non-zero entry iff an edge exists in the direct product graph and the corresponding edges in $G$ and $G'$ have the same label. Let $^lA$ denote the normalized adjacency matrix of the graph filtered by the label $l$, that is, $^lA_{ij} = A_{ij}$ if $X_{ij} = l$ and zero otherwise. Some simple algebra (omitted for the sake of brevity) shows that the weight matrix of the direct product graph can be written as

$$W_\times = \sum_{l=1}^{d} {}^lA \otimes {}^lA'. \tag{7}$$

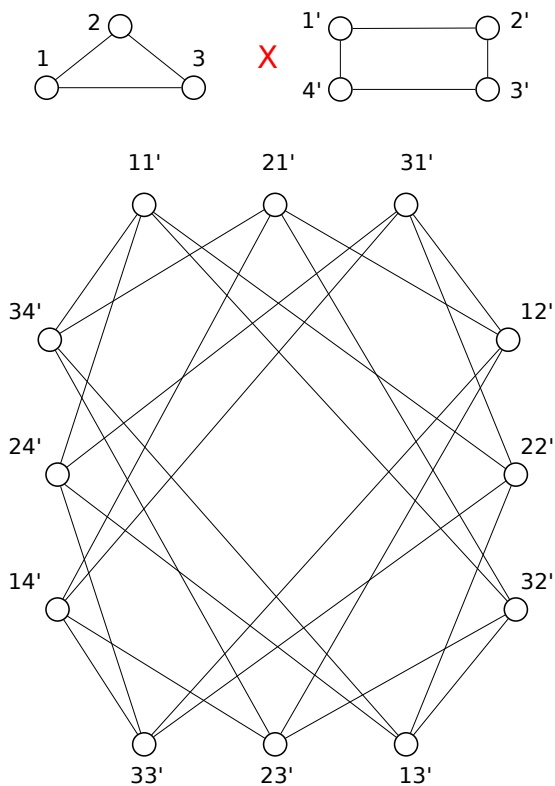

Figure 1: Two graphs (top left & right) and their direct product (bottom). Each node of the direct product graph is labeled with a pair of nodes; an edge exists in the direct product if and only if the corresponding nodes are adjacent in both original graphs. For instance, nodes $11'$ and $32'$ are adjacent because there is an edge between nodes 1 and 3 in the first, and $1'$ and $2'$ in the second graph.

We will show in the sequel that kernels defined using the above weight matrix can be computed efficiently.

## 2.2 Kernel Definition

As stated above, performing a random walk on the direct product graph $G_\times$ is equivalent to performing a simultaneous random walk on the graphs $G$ and $G'$ (Imrich and Klavžar, 2000). Therefore, the $((i-1)n+j, (i'-1)n'+j')^{\text{th}}$ entry of $A_\times^k$ represents the probability of simultaneous $k$ length random walks on $G$ (starting from vertex $v_i$ and ending in vertex $v_j$) and $G'$ (starting from vertex $v_{i'}'$ and ending in vertex $v_{j'}'$). The entries of $W_\times$ represent similarity between edges. The $((i-1)n+j, (i'-1)n'+j')^{\text{th}}$ entry of $W_\times^k$ represents the similarity between simultaneous $k$ length random walks on $G$ and $G'$ measured via the kernel function $\kappa$.

Given the weight matrix $W_\times$, initial and stopping probability distributions $p_\times$ and $q_\times$, and an appropriately chosen discrete measure $\mu$, we can define a kernel on $G$ and $G'$ as

$$k(G, G') := \sum_{k=1}^{\infty} \mu(k) \, q_\times^\top W_\times^k p_\times. \tag{8}$$

In order to show that (8) is a valid *p.s.d.* kernel we need the following technical lemma:

**Lemma 2** $\quad \forall \, k \in \mathbb{N}: \quad W_\times^k p_\times = \text{vec}[\Phi(X')^k p' \, (\Phi(X)^k p)^\top].$

**Proof** By induction over $k$. Base case: $k = 1$. Observe that

$$p_\times = (p \otimes p') \, \text{vec}(1) = \text{vec}(p' p^\top). \tag{9}$$

By using Lemma 13, $W_\times p_\times$ can be written as

$$
\begin{aligned}
[\Phi(X) \otimes \Phi(X')] \, \text{vec}(p' p^\top) &= \text{vec}[\Phi(X') p' p^\top \Phi(X)^\top] \\
&= \text{vec}[\Phi(X') p' (\Phi(X) p)^\top].
\end{aligned} \tag{10}
$$

Induction from $k$ to $k+1$: Using the induction assumption $W_\times^k p_\times = \text{vec}[\Phi(X')^k p' \, (\Phi(X)^k p)^\top]$ and Lemma 13 we obtain

$$
\begin{aligned}
W_\times^{k+1} p_\times = W_\times W_\times^k p_\times &= (\Phi(X) \otimes \Phi(X')) \, \text{vec}[\Phi(X')^k p' (\Phi(X)^k p)^\top] \\
&= \text{vec}[\Phi(X') \Phi(X')^k p' \, (\Phi(X)^k p)^\top \Phi(X)^\top] \\
&= \text{vec}[\Phi(X')^{k+1} p' \, (\Phi(X)^{k+1} p)^\top].
\end{aligned} \tag{11}
$$

∎

**Lemma 3** *If the measure $\mu(k)$ is such that* (8) *converges, then it defines a valid* p.s.d. *kernel.*

**Proof** Using Lemmas 13 and 2 we can write

$$
\begin{aligned}
q_\times^\top W_\times^k p_\times &= (q \otimes q') \, \text{vec}[\Phi(X')^k p' \, (\Phi(X)^k p)^\top] \\
&= \text{vec}[q'^\top \Phi(X')^k p' \, (\Phi(X)^k p)^\top q] \\
&= \underbrace{(q^\top \Phi(X)^k p)^\top}_{\rho(G)^\top} \underbrace{(q'^\top \Phi(X')^k p')}_{\rho(G')}.
\end{aligned} \tag{12}
$$

Each individual term of (8) equals $\rho(G)^\top \rho(G')$ for some function $\rho$, and is therefore a valid *p.s.d.* kernel. The lemma follows because the class of *p.s.d.* kernels is closed under convex combinations (Berg et al., 1984). ∎

### 2.3 Special Cases

A popular choice to ensure convergence of (8) is to assume $\mu(k) = \lambda^k$ for some $\lambda > 0$. If $\lambda$ is sufficiently small[1] then (8) is well-defined, and we can write

$$k(G, G') = \sum_k \lambda^k q_\times^\top W_\times^k p_\times = q_\times^\top (\mathbf{I} - \lambda W_\times)^{-1} p_\times. \tag{13}$$

Kashima et al. (2004) use marginalization and probabilities of random walks to define kernels on graphs. Given transition probability matrices $P$ and $P'$ associated with graphs $G$ and $G'$ respectively, their kernel can be written as (Kashima et al., 2004, Eq. 1.19)

$$k(G, G') = q_\times^\top (\mathbf{I} - T_\times)^{-1} p_\times, \tag{14}$$

where $T_\times := [\operatorname{vec}(P) \operatorname{vec}(P')^\top] \odot [\Phi(X) \otimes \Phi(X')]$. The edge kernel $\hat{\kappa}(X_{ij}, X'_{i'j'}) := P_{ij} P'_{i'j'} \kappa(X_{ij}, X'_{i,j'})$ with $\lambda = 1$ recovers (13).

Gärtner et al. (2003), on the other hand, use the adjacency matrix of the direct product graph to define the so-called random walk graph kernel

$$k(G, G') = \sum_{i=1}^n \sum_{j=1}^{n'} \sum_{k=1}^\infty \lambda^k [A_\times^k]_{ij}. \tag{15}$$

To recover their kernel in our framework, assume an uniform distribution over the vertices of $G$ and $G'$, that is, set $p = q = 1/n$ and $p' = q' = 1/n'$. The initial as well as final probability distribution over vertices of $G_\times$ is given by $p_\times = q_\times = \mathbf{e}/(nn')$. Setting $\Phi(X) := A$, $\Phi(X') = A'$, and $W_\times = A_\times$, we can rewrite (8) to obtain

$$k(G, G') = \sum_{k=1}^\infty \lambda^k q_\times^\top A_\times^k p_\times = \frac{1}{n^2 n'^2} \sum_{i=1}^n \sum_{j=1}^{n'} \sum_{k=1}^\infty \lambda^k [A_\times^k]_{ij},$$

which recovers (15) to within a constant factor.

Finally, the so-called geometric kernel is defined as (Gärtner, 2002)

$$k(G, G') = \sum_{i=1}^n \sum_{j=1}^{n'} [e^{\lambda A_\times}]_{ij} = \mathbf{e}^\top e^{\lambda A_\times} \mathbf{e}, \tag{16}$$

which can be recovered in our setting by setting $p = q = 1/n$, $p' = q' = 1/n'$, $\Phi(L) := A$, $\Phi(L') = A'$, and $\mu(k) = \lambda^k/k!$.

## 3. Efficient Computation

Computing the kernels of Gärtner et al. (2003) and Kashima et al. (2004) essentially boil down to inverting the matrix $(\mathbf{I} - \lambda W_\times)$. If both $G$ and $G'$ have $n$ vertices, then $(\mathbf{I} - \lambda W_\times)$ is an $n^2 \times n^2$ matrix. Given that the complexity of inverting a matrix is cubic in its dimensions, a direct computation of (13) would require $O(n^6)$ time. In the first part of this section we show that iterative methods, including those based on Sylvester equations, conjugate gradients, and fixed-point iterations, can be used to speed up this computation. Later, in Section 3.4, we show that the geometric kernel can be computed in $O(n^3)$ time.

---

1. The values of $\lambda$ which ensure convergence depend on the spectrum of $W_\times$. See Chapter 6 of Vishwanathan (2002) for a discussion of this issue.

### 3.1 Sylvester Equation Methods

Consider the following equation, commonly known as the Sylvester or Lyapunov equation:

$$M = SMT + M_0. \tag{17}$$

Here, $S, T, M_0 \in \mathbb{R}^{n \times n}$ are given and we need for solve for $M \in \mathbb{R}^{n \times n}$. These equations can be readily solved in $O(n^3)$ time with freely available code (Gardiner et al., 1992), such as Matlab's `dlyap` method. Solving the generalized Sylvester equation

$$M = \sum_{i=1}^{d} S_i M T_i + M_0 \tag{18}$$

involves computing generalized simultaneous Schur factorizations of $d$ symmetric matrices (Lathauwer et al., 2004). Although technically involved, this can also be solved efficiently, albeit at a higher computational cost.

We now show that if the weight matrix $W_\times$ can be written as (7) then the problem of computing the graph kernel (13) can be reduced to the problem of solving the following Sylvester equation:

$$M = \sum_{i=1}^{d} \lambda \, {}^{i}\!A' \, M \, {}^{i}\!A^\top + M_0, \tag{19}$$

where $\text{vec}(M_0) = p_\times$. We begin by *flattening* (19):

$$\text{vec}(M) = \lambda \sum_{i=1}^{d} \text{vec}({}^{i}\!A' M \, {}^{i}\!A^\top) + p_\times. \tag{20}$$

Using Lemma 13 we can rewrite (20) as

$$(\mathbf{I} - \lambda \sum_{i=1}^{d} {}^{i}\!A \otimes {}^{i}\!A') \, \text{vec}(M) = p_\times, \tag{21}$$

use (7), and solve (21) for $\text{vec}(M)$:

$$\text{vec}(M) = (\mathbf{I} - \lambda W_\times)^{-1} p_\times. \tag{22}$$

Multiplying both sides of (22) by $q_\times^\top$ yields

$$q_\times^\top \text{vec}(M) = q_\times^\top (\mathbf{I} - \lambda W_\times)^{-1} p_\times. \tag{23}$$

The right-hand side of (23) is the graph kernel (13). Given the solution $M$ of the Sylvester equation (19), the graph kernel can be obtained as $q_\times^\top \text{vec}(M)$ in $O(n^2)$ time. Since solving the Sylvester equation takes $O(n^3)$ time, computing the random walk graph kernel in this fashion is significantly faster than the $O(n^6)$ time required by the direct approach.

Solving the generalized Sylvester equation requires computing generalized simultaneous Schur factorizations of $d$ symmetric matrices, where $d$ is the number of labels. If $d$ is large, the computational cost may be reduced further by computing matrices $S$ and $T$ such that $W_\times \approx S \otimes T$. We then simply solve the simple Sylvester equation (17) involving these matrices. Finding the nearest Kronecker product approximating a matrix such as $W_\times$ is a well-studied problem in numerical linear algebra, and efficient algorithms which exploit sparsity of $W_\times$ are readily available (Van Loan, 2000). Formally, these methods minimize the Frobenius norm $||W_\times - S \otimes T||_F$ by computing the largest singular value of a permuted version of $W_\times$. In general this takes $O(n^4)$ time for an $n^2$ by $n^2$ matrix, but can be done in $O(dn^3)$ here since $W_\times$ is a sum of Kronecker products. Sparsity of $W_\times$ can then be exploited to speed this computation further.

### 3.2 Conjugate Gradient Methods

Given a matrix $M$ and a vector $b$, conjugate gradient (CG) methods solve the system of equations $Mx = b$ efficiently (Nocedal and Wright, 1999). While they are designed for symmetric *p.s.d.* matrices, CG solvers can also be used to solve other linear systems efficiently. They are particularly efficient if the matrix is rank deficient, or has a small *effective rank*, that is, number of distinct eigenvalues. Furthermore, if computing matrix-vector products is cheap — because $M$ is sparse, for instance — the CG solver can be sped up significantly (Nocedal and Wright, 1999). Specifically, if computing $Mr$ for an arbitrary vector $r$ requires $O(k)$ time, and the effective rank of the matrix is $m$, then a CG solver requires only $O(mk)$ time to solve $Mx = b$.

The graph kernel (13) can be computed by a two-step procedure: First we solve the linear system

$$(\mathbf{I} - \lambda W_\times)\, x = p_\times, \tag{24}$$

for $x$, then we compute $q_\times^\top x$. We now focus on efficient ways to solve (24) with a CG solver. Recall that if $G$ and $G'$ contain $n$ vertices each then $W_\times$ is a $n^2 \times n^2$ matrix. Directly computing the matrix-vector product $W_\times r$, requires $O(n^4)$ time. Key to our speed-ups is the ability to exploit Lemma 13 to compute this matrix-vector product more efficiently: Recall that $W_\times = \Phi(X) \otimes \Phi(X')$. Letting $r = \text{vec}(R)$, we can use Lemma 13 to write

$$W_\times r = (\Phi(X) \otimes \Phi(X'))\, \text{vec}(R) = \text{vec}(\Phi(X') R\, \Phi(X)^\top). \tag{25}$$

If $\phi(\cdot) \in \mathbb{R}^d$ then the above matrix-vector product can be computed in $O(n^3 d)$ time. If $\Phi(X)$ and $\Phi(X')$ are sparse, however, then $\Phi(X') R\, \Phi(X)^\top$ can be computed yet more efficiently: if there are $O(n)$ non-$\zeta$ entries in $\Phi(X)$ and $\Phi(X')$, then computing (25) requires only $O(n^2)$ time.

### 3.3 Fixed-Point Iterations

Fixed-point methods begin by rewriting (24) as

$$x = p_\times + \lambda W_\times x. \tag{26}$$

Now, solving for $x$ is equivalent to finding a fixed point of the above iteration (Nocedal and Wright, 1999). Letting $x_t$ denote the value of $x$ at iteration $t$, we set $x_0 := p_\times$, then

compute

$$x_{t+1} = p_\times + \lambda W_\times x_t \tag{27}$$

repeatedly until $||x_{t+1} - x_t|| < \varepsilon$, where $|| \cdot ||$ denotes the Euclidean norm and $\varepsilon$ some pre-defined tolerance. This is guaranteed to converge if all eigenvalues of $\lambda W_\times$ lie inside the unit disk; this can be ensured by setting $\lambda < 1/\xi_{\max}$, where $\xi_{\max}$ is the largest-magnitude eigenvalue of $W_\times$.

The above is closely related to the power method used to compute the largest eigenvalue of a matrix (Golub and Van Loan, 1996); efficient preconditioners can also be used to speed up convergence (Golub and Van Loan, 1996). Since each iteration of (27) involves computation of the matrix-vector product $W_\times x_t$, all speed-ups for computing the matrix-vector product discussed in Section 3.2 are applicable here. In particular, we exploit the fact that $W_\times$ is a sum of Kronecker products to reduce the worst-case time complexity to $O(n^3)$ per iteration in our experiments, in contrast to Kashima et al. (2004) who computed the matrix-vector product explicitly.

### 3.4 Geometric Kernel

We now turn our attention to the geometric kernel, (16). If both $G$ and $G'$ have $n$ vertices then $A_\times$ is a $n^2 \times n^2$ matrix, and therefore a naive implementation of the geometric kernel requires $O(n^6)$ time. The following lemma shows that this can be reduced to $O(n^3)$.

**Lemma 4** *If $G$ and $G'$ have $n$ vertices then the geometric kernel, (16), can be computed in $O(n^3)$ time.*

**Proof** Let $A = PDP^\top$ denote the spectral decomposition of $A$, that is, columns of $P$ are the eigenvectors of $A$ and $D$ is a diagonal matrix of corresponding eigenvalues (Stewart, 2000). Similarly $A' = P'D'P'^\top$. The spectral decomposition of a $n \times n$ matrix can be computed efficiently in $O(n^3)$ time (Golub and Van Loan, 1996).

Using Propositions 7.1.10, 7.1.6, and 7.1.3 of Bernstein (2005) it is easy to show that the spectral decomposition of $A_\times$ is $(P \otimes P')(D \otimes D')(P \otimes P')^\top$. Furthermore, the matrix exponential $\exp(\lambda A_\times)$ can be written as $(P \otimes P') \exp(\lambda D \otimes D')(P \otimes P')^\top$ (Bernstein, 2005, proposition 11.2.3). This and (2) allow us to rewrite (16) as

$$k(G, G') = (\mathbf{e} \otimes \mathbf{e})^\top (P \otimes P') \exp(\lambda D \otimes D')(P \otimes P')^\top (\mathbf{e} \otimes \mathbf{e}) \tag{28}$$

$$= (\mathbf{e}^\top P \otimes \mathbf{e}^\top P') \exp(\lambda D \otimes D')(P^\top \mathbf{e} \otimes P'^\top \mathbf{e}). \tag{29}$$

The proof follows by observing that each of the three terms in the above equation as well as their product can be computed in $O(n^2)$ time. ∎

## 4. Experiments

Numerous other studies have applied random walk graph kernels to applications like protein function prediction (Borgwardt et al., 2005) and chemoinformatics (Kashima et al., 2004).

Therefore we concentrate on runtime comparisons in our experimental evaluation. We present three sets of experiments. First, we work with randomly generated graphs and study the scaling behaviour of our algorithms. Second, we assess the practical impact of our algorithmic improvement by comparing the time taken to compute graph kernels on four real-world datasets whose size mandates fast kernel computation. Third, we devise novel methods for protein interaction network comparison using graph kernels. The algorithmic challenge here is to efficiently compute kernels on large sparse graphs.

For all our experiments our baseline comparator is the direct approach of Gärtner et al. (2003). Our code was written in Matlab Release 14, and experiments run under Suse Linux on a 2.6 GHz Intel Pentium 4 PC with 2 GB of main memory. We employed Lemma 13 to speed up matrix-vector multiplication for both CG and fixed-point methods (*cf.* Section 3.2), and used the function `dlyap` from the control toolbox of Matlab to solve the Sylvester equation. By default, we used a value of $\lambda = 0.001$, and set the convergence tolerance for both CG solver and fixed-point iteration to $10^{-6}$. The value of $\lambda$ was chosen to ensure that the random walk graph kernel converges. Since our methods are exact and produce the same kernel values (to numerical precision), where applicable, we only report the CPU time each algorithm takes.

### 4.1 Randomly Generated Graphs

The aim here is to study the scaling behaviour of our algorithms on graphs of different sizes and different node degrees. We generated two sets of graphs: for the first set, SET-1, we begin with an empty graph of size $2^k$, $k = 1, 2, \ldots, 10$, and randomly insert edges until a) the graph is fully connected, and b) the average degree of each node is at least 2. For each $k$ we repeat the process 10 times and generate 10 such graphs. The time required to compute the $10 \times 10$ kernel matrix for each value of $k$ is depicted in Figure 2 (left). As expected, the direct approach scales as $O(n^6)$, solving the Sylvester equation (SYL) as $O(n^3)$, while the conjugate gradient (CG) and fixed point iteration (FP) approaches scale sub-cubically. Furthermore, note that the direct approach could not handle graphs of size greater than $128 = 2^7$ even after two days of computation.

We also examined the impact of Lemma 13 on enhancing the runtime performance of the fixed-point iteration approach as originally proposed by Kashima et al. (2004). For this experiment, we again use graphs from SET-1 and computed the $10 \times 10$ kernel matrix, once using the original fixed-point iteration, and once using fixed-point iteration enhanced by Lemma 13. Results are illustrated in Figure 2 (right). Our approach is often 10 times or more faster than the original fixed-point iteration, especially on larger graphs.

The second set of randomly generated graphs is called SET-2. Here, we fixed the size of the graph at $2^5 = 32$ (the largest size that the direct method could handle comfortably), and randomly inserted edges until a) the graph is fully connected, and b) the average number of non-zero entries in the adjacency matrix is at least $x\%$, where $x = 10, 20, \ldots, 100$. For each $x$, we generate 10 such graphs and compute the $10 \times 10$ kernel matrix. Our results are shown in the left panel of Figure 3. On these small graphs the runtimes of all the methods, including the direct method, is seen to be fairly independent of the filling degree. This is not surprising since the direct method has to explicitly compute the inverse matrix; the inverse of a sparse matrix need not be sparse.

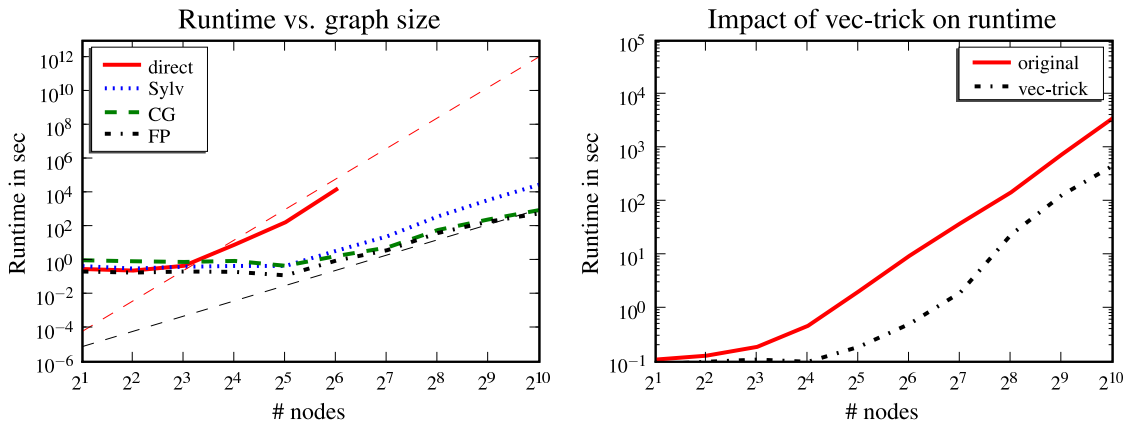

Figure 2: Time to compute a 10×10 kernel matrix on SET-1 plotted as a function of the size of graphs (# nodes). Left: We compare the Sylvester equation (SYL), conjugate gradient (CG), and fixed point iteration (FP) approaches to the direct method. The dashed thin red line indicates $O(n^6)$ scaling, while the dashed thin black line indicates $O(n^3)$ scaling. Right: We compare the runtime of the original fixed-point iteration (original) with the fixed-point iteration enhanced with Lemma 13 (vec-trick).

In order to investigate further the behavior of our speedups on large graphs we generated a new random set of graphs, by using the same procedure as for SET-2, but with the graph size to 1024. The direct method is infeasible for such large graphs. We plot the runtimes of computing the $10 \times 10$ kernel matrix in the right panel of Figure 3. On these large graphs the runtimes of the Sylvester equation solver are fairly independent of the filling degree. This is because the Sylvester equation solver is not able to exploit sparsity in the adjacency matrices. On the other hand, the runtimes of both the conjugate gradient solver as well as the fixed point iteration increase with the filling degree. Especially for very sparse graphs (filling degree of less than 20%) these methods are seen to be extremely efficient.

## 4.2 Real-World Datasets

In our next experiment we use four real-world datasets: Two sets of molecular compounds (MUTAG and PTC), and two datasets describing protein tertiary structure (Protein and Enzyme). Graph kernels provide useful measures of similarity for all of these. We now briefly describe each dataset, and discuss how graph kernels are applicable.

**Chemical Molecules** Toxicity of chemical molecules can be predicted to some degree by comparing their three-dimensional structure. We employed graph kernels to measure similarity between molecules from the MUTAG and PTC datasets (Toivonen et al., 2003). The average number of nodes per graph in these sets is 17.72 *resp.* 26.70; the average number of edges is 38.76 *resp.* 52.06.

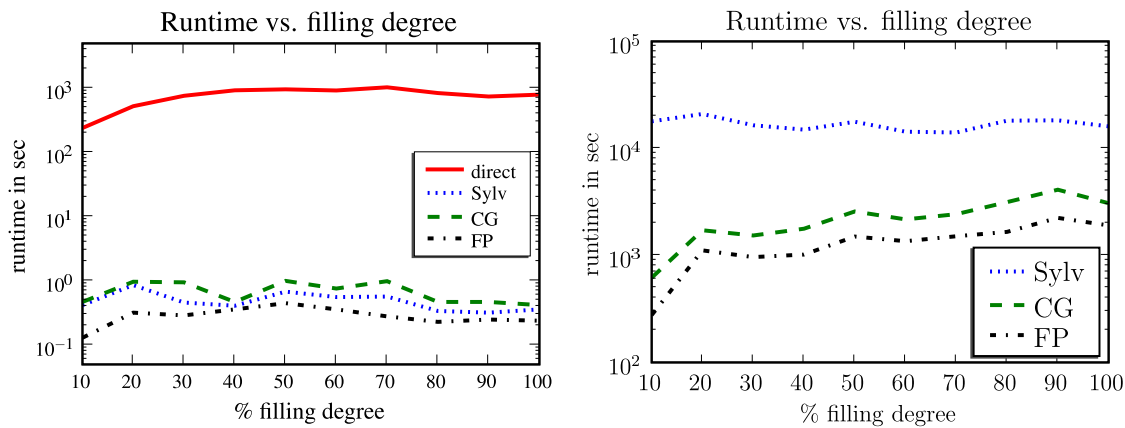

Figure 3: Time to compute a $10 \times 10$ kernel matrix on SET-1 plotted as a function of the filling degree of the graph. Left: We compare the Sylvester equation (SYL), conjugate gradient (CG), and fixed point iteration (FP) approaches to the direct method on graphs containing 32 nodes. Right: We compare SYL, CG, and FP approaches on larger graphs with 1024 nodes. The direct method is infeasible on these graphs.

**Large Protein Graph Dataset**    A standard approach to protein function prediction entails classifying proteins into enzymes and non-enzymes, then further assigning enzymes to one of the six top-level classes of the Enzyme Commission (EC) hierarchy. Towards this end, Borgwardt et al. (2005) modeled a dataset of 1128 proteins as graphs in which vertices represent secondary structure elements, and edges represent neighborhood within the 3-D structure or along the amino acid chain. Comparing these graphs via a modified random walk graph kernel and classifying them via a Support Vector Machine (SVM) led to function prediction accuracies competitive with state-of-the-art approaches (Borgwardt et al., 2005). We used Borgwardt et al.'s (2005) data to test the efficacy of our methods on a large dataset. The average number of nodes and edges per graph in this data is 38.57 *resp.* 143.75. We used a single label on the edges, and the delta kernel to define similarity between edges.

**Large Enzyme Graph Dataset**    We repeated the above experiment on an enzyme graph dataset, also due to Borgwardt et al. (2005). This dataset contains 600 graphs, with 32.63 nodes and 124.27 edges on average. Graphs in this dataset represent enzymes from the BRENDA enzyme database (Schomburg et al., 2004). The biological challenge on this data is to correctly assign the enzymes to one of the EC top-level classes.

### 4.2.1 UNLABELED GRAPHS

For this experiment, we computed kernels taking into account only the topology of the graph, *i.e.*, we did not consider node or edge labels. Table 1 lists the CPU time required to compute the full kernel matrix for each dataset, as well as — for comparison purposes — a $100 \times 100$ sub-matrix. The latter is also shown graphically in Figure 4 (left).

Table 1: Time to compute kernel matrix for unlabeled graphs from various datasets.

| dataset | MUTAG | | PTC | | Enzyme | | Protein | |
|---|---|---|---|---|---|---|---|---|
| nodes/graph | 17.7 | | 26.7 | | 32.6 | | 38.6 | |
| edges/node | 2.2 | | 1.9 | | 3.8 | | 3.7 | |
| #graphs | 100 | 230 | 100 | 417 | 100 | 600 | 100 | 1128 |
| Direct | 18'09" | 104'31" | 142'53" | 41h* | 31h* | 46.5d* | 36d* | 12.5y* |
| Sylvester | 25.9" | 2'16" | 73.8" | 19'30" | 48.3" | 36'43" | 69'15" | 6.1d* |
| Conjugate | 42.1" | 4'04" | 58.4" | 19'27" | 44.6" | 34'58" | 55.3" | 97'13" |
| Fixed-Point | 12.3" | 1'09" | 32.4" | 5'59" | 13.6" | 15'23" | 31.1" | 40'58" |

∗: Extrapolated; run did not finish in time available.

On these unlabeled graphs, conjugate gradient and fixed-point iteration — sped up via our Lemma 13 — are consistently about two orders of magnitude faster than the conventional direct method. The Sylvester equation approach is very competitive on smaller graphs (outperforming CG on MUTAG) but slows down with increasing number of nodes per graph; this is because we were unable to incorporate Lemma 13 into Matlab's black-box `dlyap` solver. Even so, the Sylvester equation approach still greatly outperforms the direct method.

### 4.2.2 Labeled Graphs

For this experiment, we compared graphs with edge labels. Note that node labels can be dealt with by concatenating them to the edge labels of adjacent edges. On the two protein datasets we employed a linear kernel to measure similarity between edge labels representing distances (in Ångströms) between secondary structure elements. On the two chemical datasets we used a delta kernel to compare edge labels reflecting types of bonds in molecules. We report CPU times for the full kernel matrix as well as a $100 \times 100$ sub-matrix in Table 2; the latter is also shown graphically in Figure 4 (right).

On labeled graphs, our three methods outperform the direct approach by about a factor of 1000 when using the linear kernel. In the experiments with the delta kernel, conjugate gradient and fixed-point iteration are still at least two orders of magnitude faster. Since we did not have access to a generalized Sylvester equation (18) solver, we had to use a Kronecker product approximation (Van Loan, 2000) which dramatically slowed down the Sylvester equation approach for the delta kernel.

## 4.3 Protein Interaction Networks

In our third experiment, we used random walk graph kernels to tackle a large-scale problem in bioinformatics involving the comparison of fairly large protein-protein interaction (PPI) networks. Using a combination of human PPI and clinical microarray gene expression data, the task is to predict the disease outcome (dead or alive, relapse or no relapse) of cancer patients. As before, we set $\lambda = 0.001$ and the convergence tolerance to $10^{-6}$ for all our experiments reported below.

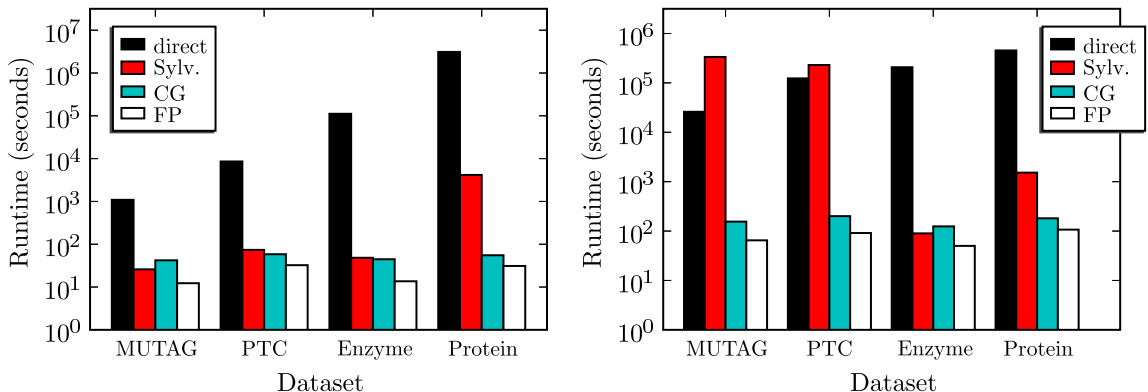

Figure 4: Time (in seconds on a log-scale) to compute 100×100 kernel matrix for unlabeled (left) *resp.* labelled (right) graphs from several datasets, comparing the conventional direct method to our fast Sylvester equation, conjugate gradient (CG), and fixed-point iteration (FP) approaches.

### 4.3.1 CO-INTEGRATION OF GENE EXPRESSION AND PPI DATA

We co-integrated clinical microarray gene expression data for cancer patients with known human PPI due to [Rual et al. (2005)](#). Specifically, a patient's gene expression profile was transformed into a graph as follows: A node was created for every protein which — according to [Rual et al. (2005)](#) — participates in an interaction, and whose corresponding gene expression level was measured on this patient's microarray. We connect two proteins in this graph by an edge if [Rual et al. (2005)](#) list these proteins as interacting, and both genes are up- *resp.* downregulated with respect to a reference measurement. Each node bears the name of the corresponding protein as its label.

Table 2: Time to compute kernel matrix for labeled graphs from various datasets.

| kernel | delta | | | | linear | | | |
|---|---|---|---|---|---|---|---|---|
| dataset | MUTAG | | PTC | | Enzyme | | Protein | |
| #graphs | 100 | 230 | 100 | 417 | 100 | 600 | 100 | 1128 |
| Direct | 7.2h | 1.6d* | 1.4d* | 25d* | 2.4d* | 86d* | 5.3d* | 18y* |
| Sylvester | 3.9d* | 21d* | 2.7d* | 46d* | 89.8" | 53'55" | 25'24" | 2.3d* |
| Conjugate | 2'35" | 13'46" | 3'20" | 53'31" | 124.4" | 71'28" | 3'01" | 4.1h |
| Fixed-Point | 1'05" | 6'09" | 1'31" | 26'52" | 50.1" | 35'24" | 1'47" | 1.9h |

∗: Extrapolated; run did not finish in time available.

This approach of co-integrating PPI and gene expression data is built on the assumption that genes with similar gene expression levels are translated into proteins that are more likely to interact. Recent studies confirm that this assumption holds significantly more often for co-expressed than for random pairs of proteins (Fraser et al., 2004; Bhardwaj and Lu, 2005). To measure similarity between these networks in a biologically meaningful manner, we compare which groups of proteins interact and are co-regulated in each patient. For this purpose, a random walk graph kernel is the natural choice of kernel, as a random walk in this graph represents a group of proteins, in which consecutive proteins along the walk are co-expressed and interact. As each node bears the name of its corresponding protein as its node label, the size of the product graph is at most that of the smaller of the two input graphs.

### 4.3.2 Composite Graph Kernel

The presence of an edge in a graph signifies an interaction between the corresponding nodes. In chemoinformatics, for instance, edges indicate chemical bonds between two atoms; in PPI networks, edges indicate interactions between proteins. When studying protein interactions in disease, however, the *absence* of a given interaction can be as significant as its presence. Since existing graph kernels cannot take this into account, we propose to modify them appropriately. Key to our approach is the notion of a complement graph:

**Definition 5** *Let $G = (V, E)$ be a graph with vertex set $V$ and edge set $E$. Its complement $\bar{G} = (V, \bar{E})$ is a graph over the same vertices but with complementary edges $\bar{E} := (V \times V) \backslash E$.*

In other words, the complement graph consists of exactly those edges *not* present in the original graph. Using this notion we define the *composite* graph kernel

$$k_{comp}(G, G') := k(G, G') + k(\bar{G}, \bar{G}').\tag{30}$$

This deceptively simple kernel leads to substantial gains in performance in our experiments comparing co-integrated gene expression/protein interaction networks.

### 4.3.3 Datasets

**Leukemia.** Bullinger et al. (2004) provide a dataset of microarrays of 119 leukemia patients. Since 50 patients survived after a median follow-up time of 334 days, always predicting a lethal outcome here would result in a baseline prediction accuracy of 1 - 50/119 = 57.98%. Co-integrating this data with human PPI, we found 2, 167 proteins from Rual et al. (2005) for which Bullinger et al. (2004) report expression levels among the 26, 260 genes they examined.

**Breast Cancer.** This dataset consists of microarrays of 78 breast cancer patients, of which 44 had shown no relapse of metastases within 5 years after initial treatment (van't Veer et al., 2002). Always predicting survival thus gives a baseline prediction accuracy of 44/78 = 56.41% on this data. When generating co-integrated graphs, we found 2, 429 proteins from Rual et al. (2005) for which van't Veer et al. (2002) measure gene expression out of the 24, 479 genes they studied.

Table 3: Average time to compute kernel matrix on protein interaction networks.

| dataset | Leukemia | | Breast Cancer | |
|---|---|---|---|---|
| kernel | vanilla | composite | vanilla | composite |
| Direct | 2h 15'23" | 5h 12'29" | 4h 01'16" | 8h 24'45" |
| Sylvester | 12'03" | 25'41" | 20'21" | 45'51" |
| Conjugate | 6" | 13" | 13" | 28" |
| Fixed-Point | 4" | 7" | 8" | 17" |

### 4.3.4 Results

The CPU runtimes of our CG, fixed-point, and Sylvester equation approaches to graph kernel computation on the cancer patients modeled as graphs is contrasted with that of the direct approach in Table 3. Using the computed kernel and a support vector machine (SVM) we tried to predict the survivors, either with a "vanilla" graph kernel (13), or our composite graph kernel (30) in 10-fold cross-validation.

On both datasets, our approaches to fast graph kernel computation convey up to three orders of magnitude gain in speed. With respect to prediction accuracy, the vanilla random walk graph kernel performs hardly better than the baseline classifer on both tasks (Leukemia: 59.17 % vs 57.98 %; Breast Cancer: 56.41 % vs. 56.41 %). The composite graph kernel outperforms the vanilla graph kernel in accuracy in both experiments, with an increase in prediction accuracy of around 4–5 % (Leukemia: 63.33 %; Breast cancer: 61.54 %).

The vanilla kernel suffers from its inability to measure network discrepancies, the paucity of the graph model employed, and the fact that only a small minority of genes could be mapped to interacting proteins; due to these problems, its accuracy remains close to the baseline. The composite kernel, by contrast, also models missing interactions. With it, even our simple graph model, that only captures 10% of the genes examined in both studies, is able to capture some relevant biological information, which in turn leads to better classification accuracy on these challenging datasets (Warnat et al., 2005).

## 5. Rational Kernels

The aim of this section is to establish connections between rational kernels on transducers (Cortes et al., 2004) and random walk graph kernels. In particular, we show that composition of transducers is analogous to computing product graphs, and that rational kernels on weighted transducers may be viewed as generalizations of random walk graph kernels to weighted automata. In order to make these connections explicit we adapt slightly non-standard notation for weighted transducers, extensively using matrices and tensors wherever possible.

### 5.1 Semiring

At the most general level, weighted transducers are defined over semirings. In a semiring addition and multiplication are generalized to abstract operations $\bar{\oplus}$ and $\bar{\odot}$ with the same distributive properties:

**Definition 6 (Mohri, 2002)** *A semiring is a system* $(\mathbb{K}, \bar{\oplus}, \bar{\odot}, \bar{0}, \bar{1})$ *such that*

1. $(\mathbb{K}, \bar{\oplus}, \bar{0})$ *is a commutative monoid with* $\bar{0}$ *as the identity element for* $\bar{\oplus}$ *(i.e., for any* $x, y, z \in \mathbb{K}$*, we have* $x \bar{\oplus} y \in \mathbb{K}$*,* $(x \bar{\oplus} y) \bar{\oplus} z = x \bar{\oplus} (y \bar{\oplus} z)$*,* $x \bar{\oplus} \bar{0} = \bar{0} \bar{\oplus} x = x$ *and* $x \bar{\oplus} y = y \bar{\oplus} x$*);*

2. $(\mathbb{K}, \bar{\odot}, \bar{1})$ *is a monoid with* $\bar{1}$ *as the identity operator for* $\bar{\odot}$ *(i.e., for any* $x, y, z \in \mathbb{K}$*, we have* $x \bar{\odot} y \in \mathbb{K}$*,* $(x \bar{\odot} y) \bar{\odot} z = x \bar{\odot} (y \bar{\odot} z)$*, and* $x \bar{\odot} \bar{1} = \bar{1} \bar{\odot} x = x$*);*

3. $\bar{\odot}$ *distributes over* $\bar{\oplus}$*, i.e., for any* $x, y, z \in \mathbb{K}$*,*

$$(x \bar{\oplus} y) \bar{\odot} z = (x \bar{\odot} z) \bar{\oplus} (y \bar{\odot} z), \tag{31}$$

$$z \bar{\odot} (x \bar{\oplus} y) = (z \bar{\odot} x) \bar{\oplus} (z \bar{\odot} y); \tag{32}$$

4. $\bar{0}$ *is an annihilator for* $\bar{\odot}$*:* $\forall x \in \mathbb{K}, \ x \bar{\odot} \bar{0} = \bar{0} \bar{\odot} x = \bar{0}$*.*

Thus, a semiring is a ring that may lack negation. $(\mathbb{R}, +, \cdot, 0, 1)$ is the familiar semiring of real numbers. Other examples include

**Boolean:** $(\{\text{FALSE}, \text{TRUE}\}, \vee, \wedge, \text{FALSE}, \text{TRUE})$;

**Logarithmic:** $(\mathbb{R} \cup \{-\infty, \infty\}, \bar{\oplus}_{\ln}, +, -\infty, 0)$, where $\forall x, y \in \mathbb{K}: \ x \bar{\oplus}_{\ln} y := \ln(e^x + e^y)$;

**Tropical:** $(\mathbb{R}^+ \cup \{-\infty\}, \max, +, -\infty, 0)$.

The $(\bar{\oplus}, \bar{\odot})$ operations in some semirings can be mapped into ordinary $(+, \cdot)$ operations by applying an appropriate morphism.

**Definition 7** *Let* $(\mathbb{K}, \bar{\oplus}, \bar{\odot}, \bar{0}, \bar{1})$ *be a semiring. A function* $\psi : \mathbb{K} \to \mathbb{R}$ *is a morphism if*

$$\psi(x \bar{\oplus} y) = \psi(x) + \psi(y); \tag{33}$$

$$\psi(x \bar{\odot} y) = \psi(x) \cdot \psi(y); \tag{34}$$

$$\psi(\bar{0}) = 0 \ \text{and} \ \psi(\bar{1}) = 1. \tag{35}$$

In the following, by 'morphism' we will always mean a morphism from a semiring to the real numbers. Not all semirings have such morphisms. For instance, the logarithmic semiring has a morphism — namely, the exponential function — but the tropical semiring does not have one.

Ordinary linear algebra operations including Kronecker products, matrix addition, matrix-vector multiplication, and matrix-matrix multiplication can be carried over to a semiring in a straightforward manner. For instance, if $A, B \in \mathbb{K}^{n \times n}$, and $x \in \mathbb{K}^n$ then

$$[A \bar{\odot} x]_i = \bigoplus_{j=1}^{n} A_{ij} \bar{\odot} x_j, \ \text{and} \tag{36}$$

$$[A \bar{\odot} B]_{i,j} = \bigoplus_{k=1}^{n} A_{ik} \bar{\odot} B_{kj}. \tag{37}$$

As in Appendix A, we can extend a morphism $\psi$ to matrices (and analogously to vectors) by defining $[\Psi(A)]_{ij} := \psi(A_{ij})$, and $[\Psi^{-1}(A)]_{ij} := \psi^{-1}(A_{ij})$. If the semiring has a morphism, $\psi$, then it is easy to see that

$$[A \bar{\odot} x]_i = \sum_{j=1}^{n} \psi(A_{ij}) \cdot \psi(x_j) = \Psi^{-1}(\Psi(A)\Psi(x)), \text{ and} \tag{38}$$

$$[A \bar{\odot} B]_{i,j} = \sum_{k=1}^{n} A_{ik} \cdot B_{kj} = \Psi^{-1}(\Psi(A)\Psi(B)). \tag{39}$$

## 5.2 Weighted Transducers

Loosely speaking, a transducer is a weighted automaton with an input and output alphabet. We will work with the following slightly specialized definition[2]:

**Definition 8** *A weighted finite-state transducer $T$ over a semiring $(\mathbb{K}, \bar{\oplus}, \bar{\odot}, \bar{0}, \bar{1})$ is a 5-tuple $T = (\Sigma, Q, A, p, q)$, where $\Sigma$ is a finite input-output alphabet, $Q$ is a finite set of $n$ states, $p \in \mathbb{K}^n$ is a vector of initial weights, $q \in \mathbb{K}^n$ is a vector of final weights, and $A$ is a 4-dimensional tensor in $\mathbb{K}^{n \times |\Sigma| \times |\Sigma| \times n}$ which encodes transitions and their corresponding weights.*

For $a, b \in \Sigma$ we will use the shorthand $A_{ab}$ to denote the $n \times n$ slice $A_{*ab*}$ of the transition tensor, which represents all valid transitions on input label $a$ emitting the output label $b$. The output weight associated by $T$ to a pair of strings $\alpha = a_1 a_2 \ldots a_l$ and $\beta = b_1 b_2 \ldots b_l$ is

$$[\![T]\!](\alpha, \beta) = q^\top \bar{\odot} A_{a_1 b_1} \bar{\odot} A_{a_2 b_2} \bar{\odot} \ldots \bar{\odot} A_{a_l b_l} \bar{\odot} p. \tag{40}$$

A transducer is said to accept a pair of strings $(\alpha, \beta)$, if it assigns non-zero output weight to them, *i.e.*, $[\![T]\!](\alpha, \beta) \neq \bar{0}$. A transducer is said to be regulated if the output weight associated to any pair of strings is well defined in $\mathbb{K}$. Since we disallow $\epsilon$ transitions, our transducers are always regulated.

A weighted automaton is a transducer with identical input and output labels. Therefore, the transition matrix of a weighted automaton is a 3-dimensional tensor in $\mathbb{K}^{n \times |\Sigma| \times n}$. A graph is a weighted automaton whose input-output alphabet contains exactly one label, and therefore it only accepts strings of the form $a^k = aa \ldots a$. The transition matrix of a graph (equivalently, its adjacency matrix) is a 2-dimensional tensor in $\mathbb{K}^{n \times n}$. If $A$ denotes the adjacency matrix of a graph $G$, then the output weight assigned by $G$ to $a^k$ is

$$[\![G]\!](a^k) = q^\top \bar{\odot} A \bar{\odot} A \bar{\odot} \ldots \bar{\odot} A \bar{\odot} p. \tag{41}$$

The inverse of $T = (\Sigma, Q, A, p, q)$, denoted by $T^{-1}$, is obtained by transposing the input and output labels of each transition. Formally, $T^{-1} = (\Sigma, Q, B, p, q)$ where $B_{ab} = A_{ba}$.

The composition of two automata $T = (\Sigma, Q, A, p, q)$ and $T' = (\Sigma, Q', A', p', q')$ is an automaton $T_\times = T \circ T' = (\Sigma, Q_\times, B, p_\times, q_\times)$, where $Q_\times = Q \times Q'$, $p_\times = p \bar{\otimes} p'$[3], $q_\times := q \bar{\otimes} q'$, and $B_{ab} = \bigoplus_{c \in \Sigma} A_{ac} \bar{\otimes} A'_{cb}$. In particular, composing $T$ with its inverse yields $T \circ T^{-1} = (\Sigma, Q \times Q, B, p \bar{\otimes} p, q \bar{\otimes} q)$, where $B_{ab} = \bigoplus_{c \in \Sigma} A_{ac} \bar{\otimes} A_{bc}$. There exists a general and efficient algorithm for composing transducers which takes advantage of the sparseness of the input transducers (e.g. [Mohri et al., 1996](#); [Pereira and Riley, 1997](#)). Note that the composition operation, when specialized to graphs, is equivalent to computing a direct product graph.

---

2. We disallow $\epsilon$ transitions, and use the same alphabet for both input and output. Furthermore, in a departure from tradition, we represent the transition function as a 4-dimensional tensor.

3. We use $\bar{\otimes}$ to denote the Kronecker product using the semiring operation $\bar{\odot}$, in order to distinguish it from the regular Kronecker product $\otimes$.

## 5.3 Kernel Definition

Given a weighted transducer $T$ and a function $\psi : \mathbb{K} \to \mathbb{R}$, the rational kernel between two strings $\alpha = a_1 a_2 \ldots a_l$ and $\beta = b_1 b_2 \ldots b_l$ is defined as ([Cortes et al., 2004](#)):

$$k(\alpha, \beta) = \psi \left( [\![T]\!] (\alpha, \beta) \right). \tag{42}$$

[Cortes et al. (2004)](#) show that a generic way to obtain *p.s.d.* rational kernels is to replace $T$ by $T \circ T^{-1}$, and let $\psi$ be a semiring morphism. We now present an alternate proof which uses properties of the Kronecker product. Since $\psi$ is a semiring morphism, by specializing [(40)](#) to $T \circ T^{-1}$, we can write $k(\alpha, \beta) = \psi \left( [\![T \circ T^{-1}]\!] (\alpha, \beta) \right)$ as

$$\Psi(q \bar{\otimes} q)^\top \Psi \left( \bigoplus_{c_1} A_{a_1 c_1} \bar{\otimes} A_{b_1 c_1} \right) \ldots \Psi \left( \bigoplus_{c_l} A_{a_l c_l} \bar{\otimes} A_{b_l c_l} \right) \Psi(p \bar{\otimes} p), \tag{43}$$

which, in turn, can be rewritten using

$$\Psi \left( \bigoplus_{c \in \Sigma} A_{ac} \bar{\otimes} A_{bc} \right) = \sum_{c \in \Sigma} \Psi(A_{ac}) \otimes \Psi(A_{bc}) \tag{44}$$

as

$$\sum_{c_1 c_2 \ldots c_l} \Psi(q)^\top \otimes \Psi(q)^\top \left( \Psi(A_{a_1 c_1}) \otimes \Psi(A_{b_1 c_1}) \right) \ldots \left( \Psi(A_{a_l c_l}) \otimes \Psi(A_{b_l c_l}) \right) \Psi(p) \otimes \Psi(p). \tag{45}$$

By successively applying [(2)](#) we obtain

$$k(\alpha, \beta) = \sum_{c_1 c_2 \ldots c_l} \underbrace{\left( \Psi(q)^\top \Psi(A_{a_1 c_1}) \ldots \Psi(A_{a_l c_l}) \Psi(p) \right)}_{\rho(\alpha)} \underbrace{\left( \Psi(q)^\top \Psi(A_{b_1 c_1}) \ldots \Psi(A_{b_l c_l}) \Psi(p) \right)}_{\rho(\beta)}, \tag{46}$$

which shows that each individual term in the summation is a valid *p.s.d.* kernel. Since *p.s.d.* kernels are closed under addition, $k(\alpha, \beta)$ is a valid *p.s.d.* kernel.

## 5.4 Kernels on Weighted Transducers

Rational kernels on strings can be naturally extended to weighted transducers $S$ and $U$ via ([Cortes et al., 2004](#)):

$$k(S, U) = \psi \left( \bigoplus_{\alpha, \beta} [\![S]\!] (\alpha) \bar{\odot} [\![T]\!] (\alpha, \beta) \bar{\odot} [\![U]\!] (\beta) \right), \tag{47}$$

which, in turn, can be rewritten as

$$k(S, U) = \psi \left( \bigoplus_{\alpha, \beta} [\![ S \circ T \circ U ]\!] \, (\alpha, \beta) \right). \tag{48}$$

If $\psi$ is a semiring morphism, then

$$k(S, U) = \sum_{\alpha, \beta} \psi \left( [\![ S \circ T \circ U ]\!] \, (\alpha, \beta) \right). \tag{49}$$

Since *p.s.d.* kernels are closed under addition, if $\psi \left( [\![ S \circ T \circ U ]\!] \, (\alpha, \beta) \right)$ is a *p.s.d.* kernel, then $k(S, U)$ is also a valid *p.s.d.* kernel.

## 5.5 Recovering Random Walk Graph Kernels

In order to recover random walk graph kernels we use the standard $(\mathbb{R}, +, \cdot, 0, 1)$ ring as our semiring, and hence set $\psi$ to be the identity function. Note that since we are dealing with graphs, the only strings which are assigned non-zero weight are of the form $a^k = aa \ldots a$. Finally, we set the transducer $T$ to simply accept all strings of the form $a^k$ with unit weight. In this case, the kernel specializes to

$$k(G, G') = \sum_{a^k} [\![ G \circ G' ]\!] \, (a^k). \tag{50}$$

Recall that the normalized adjacency matrix of $G \circ G'$ is $A_\times := A \otimes A'$, where $A$ and $A'$ are the normalized adjacency matrices of $G$ and $G'$ respectively. By specializing (41) to $G \circ G'$ we can rewrite (50) as

$$k(G, G') = \sum_k q_\times A_\times^k p_\times. \tag{51}$$

This essentially recovers (8) with the weight matrix set to the adjacency matrix, but, without the discrete measure $\mu(k)$.

## 6. R-convolution Kernels

Haussler's (1999) R-convolution kernels provide a generic way to construct kernels for discrete compound objects. Let $x \in \mathcal{X}$ be such an object, and $\boldsymbol{x} := (x_1, x_2, \ldots, x_D)$ denote a decomposition of $x$, with each $x_i \in \mathcal{X}_i$. We can define a boolean predicate

$$R : \mathcal{X} \times \boldsymbol{\mathcal{X}} \rightarrow \{\text{TRUE}, \text{FALSE}\}, \tag{52}$$

where $\boldsymbol{\mathcal{X}} := \mathcal{X}_1 \times \ldots \times \mathcal{X}_D$ and $R(x, \boldsymbol{x})$ is TRUE whenever $\boldsymbol{x}$ is a valid decomposition of $x$. This allows us to consider the set of all valid decompositions of an object:

$$R^{-1}(x) := \{\boldsymbol{x} | R(x, \boldsymbol{x}) = \text{TRUE}\}. \tag{53}$$

Like Haussler (1999) we assume that $R^{-1}(x)$ is countable. We define the R-convolution $\star$ of the kernels $\kappa_1, \kappa_2, \ldots, \kappa_D$ with $\kappa_i : \mathcal{X}_i \times \mathcal{X}_i \to \mathbb{R}$ to be

$$k(x, x') = \kappa_1 \star \kappa_2 \star \ldots \star \kappa_D(x, x') := \sum_{\substack{\boldsymbol{x} \in R^{-1}(x) \\ \boldsymbol{x}' \in R^{-1}(x')}} \mu(\boldsymbol{x}, \boldsymbol{x}') \prod_{i=1}^{D} \kappa_i(x_i, x_i'), \qquad (54)$$

where $\mu$ is a finite measure on $\mathcal{X} \times \mathcal{X}$ which ensures that the above sum converges.[4] Haussler (1999) showed that $k(x, x')$ is *p.s.d.* and hence admissible as a kernel (Schölkopf and Smola, 2002), provided that all the individual $\kappa_i$ are. The deliberate vagueness of this setup in regard to the nature of the underlying decomposition leads to a rich framework: Many different kernels can be obtained by simply changing the decomposition.

## 6.1 Graph Kernels as R-Convolutions

To apply R-convolution kernels to graphs, one decomposes the graph into smaller substructures, and builds the kernel based on similarities between those components. Most graph kernels are—knowingly or not—based on R-convolutions; they mainly differ in the way they decompose the graph for comparison and the similarity measure they use to compare the components.

Random walk graph kernels, as proposed by Gärtner et al. (2003), decompose a graph into paths and compute a delta kernel between nodes. Borgwardt et al. (2005), on the other hand, use a kernel defined on nodes and edges in order to compute similarity between random walks. As we saw in Section 2.3, the marginalized graph kernels of Kashima et al. (2004) are closely related, if motivated differently. The decomposition corresponding to this kernel is the set of all possible label sequences generated by a walk on the graph. Mahé et al. (2004) extend this approach in two ways: They enrich the labels via the so-called Morgan index, and modify the kernel definition to prevent *tottering*, that is, the generation of high similarity scores by multiple, similar, small substructures. Both these extensions are particularly relevant for chemoinformatics applications.

Horvath et al. (2004) decompose a graph into cyclic patterns, then count the number of common cyclic patterns which occur in both graphs. Their kernel is plagued by computational issues; in fact they show that computing the cyclic pattern kernel on a general graph is NP-hard. They consequently restrict their attention to practical problem classes where the number of simple cycles is bounded.

Other decompositions of graphs, which are well suited for particular application domains, include subtrees (Ramon and Gärtner, 2003), shortest paths (Borgwardt and Kriegel, 2005), molecular fingerprints based on various types of depth-first searches (Ralaivola et al., 2005), and structural elements such as rings, functional groups (Fröhlich et al., 2006), and so on.

---

4. Haussler (1999) implicitly assumed this sum to be well-defined, and hence did not use a measure $\mu$ in his definition.

### 6.2 R-Convolutions in Abstract Semirings

There have been a few attempts to extend the R-convolution kernel (54) to abstract semirings, by defining:

$$k(x, x') := \bigoplus_{\substack{\boldsymbol{x} \in R^{-1}(x) \\ \boldsymbol{x}' \in R^{-1}(x')}} \mu(\boldsymbol{x}, \boldsymbol{x}') \odot \bigodot_{i=1}^{D} \kappa_i(x_i, x'_i). \tag{55}$$

The optimal assignment graph kernel of Fröhlich et al. (2006) is motivated along these lines, using the tropical semiring. It is defined as

$$k(x, x') = \max_{\substack{\boldsymbol{x} \in R^{-1}(x) \\ \boldsymbol{x}' \in R^{-1}(x')}} \mu(\boldsymbol{x}, \boldsymbol{x}') \sum_{i=1}^{D} \kappa_i(x_i, x'_i). \tag{56}$$

Unfortunately this kernel is not always *p.s.d.* (Vert, 2008). Establishing necessary and sufficient conditions for (55) to be *p.s.d.* remains an open problem.

## 7. Diffusion-Based Graph Kernels?

The adjacency matrix and its normalized cousin are not the only $n \times n$ matrices associated with undirected graphs. Spectral graph theorist instead prefer to use the so-called graph Laplacian

$$\widetilde{\mathrm{L}}_{ij} = [D - \widetilde{\mathrm{A}}]_{ij} = \begin{cases} -w_{ij} & \text{if } i \sim j \\ d_i & \text{if } i = j \\ 0 & \text{otherwise} \end{cases} \tag{57}$$

or the normalized Laplacian $L = D^{-1/2} \widetilde{\mathrm{L}} D^{-1/2}$. One can extend the concept of a feature matrix of a graph, $\Phi(X)$, to the Laplacian: set $\Phi(D)$ to be a diagonal matrix with diagonal entries $[\Phi(D)]_{ii} = \sum_j [\Phi(X)]_{ij}$ and non-diagonal entries $\zeta$ (the null label). Now define $\Phi(L) := \Phi(D) - \Phi(X)$.

Just as $\widetilde{\mathrm{A}}$ is related to random walks, $\widetilde{\mathrm{L}}$ is closely connected to the concept of diffusion. In fact, diffusion can be regarded as the continuous time limit of a specific type of random walk, in which over each infinitesimal time interval of length $\epsilon$, a particle at node $v_i$ will either move to one of its neighbors $v_j$ with probability $\epsilon w_{ij}$, or stay at $v_i$ with probability $1 - \epsilon \sum_{i \sim j} w_{ij} = 1 - \epsilon d_i$. Setting $\epsilon = 1/m$ for some integer $m$ going to infinity, it is easy to see that for any finite time interval of length $t$ the transition matrix of this process is

$$K_t = \lim_{m \to \infty} \left( I - \frac{t \widetilde{\mathrm{L}}}{m} \right)^m =: \exp(t \widetilde{\mathrm{L}}), \tag{58}$$

the matrix exponential of $t \widetilde{\mathrm{L}}$. The ability of the random walk to stay in place is crucial to taking the continuous time limit.

### 7.1 Diffusion Kernels on Graph Vertices

Note that in contrast to the normalized adjacency matrix $A$, the Laplacian is a symmetric matrix. Exploiting this property and the fact that (58) is unchanged if we let $m = 2m'$ and now make $m'$ go to infinity (*i.e.*, we force $m$ to be even), we may equivalently write

$$K_t = \exp(t\,\widetilde{\mathsf{L}}) = \lim_{m' \to \infty} \left[ \left( I - \frac{t\,\widetilde{\mathsf{L}}}{2m'} \right)^{m'} \right]^{\top} \left( I - \frac{t\,\widetilde{\mathsf{L}}}{m} \right)^{m'}. \tag{59}$$

Since the product of any matrix with its transpose is *p.s.d.*, we conclude that $K_t$ is symmetric and *p.s.d.*, and thus $k(v_i, v_j) := [K_t]_{ij}$ is a valid candidate for a kernel between graph vertices (Kondor and Lafferty, 2002).

The justification why $K_t$ should be a good kernel for learning problems comes from deeper arguments connecting spectral graph theory and regularization (Smola and Kondor, 2003). For example, given any function $f \colon V \to \mathbb{R}$, and letting $\mathbf{f} = (f(v_1), f(v_2), \ldots, f(v_n))^{\top}$, it is easy to see that

$$\mathbf{f}^{\top}\,\widetilde{\mathsf{L}}\,\mathbf{f} = \sum_{v_i \sim v_j} w_{ij}\,[f(v_i) - f(v_j)]^2. \tag{60}$$

The right-hand side of this equation is a natural measure of the variation of $f$ across edges, showing that in some well defined sense the Laplacian captures the smoothness of functions on graphs. In fact, it can be shown that the regularization scheme implied by the diffusion kernel is just the discretized, graph-adapted cousin of the regularization theory behind the familiar Gaussian kernel (Smola and Kondor, 2003).

In applications it often makes sense to modify the above picture somewhat and instead of $\widetilde{\mathsf{L}}$ use the normalized Laplacian $L$, mostly because of an important result stating that the eigenvalues of the latter are bounded between 0 and 2 (Chung-Graham, 1997). Breaking away from the original diffusion interpretation, but still adhering to the spectral graph theory dogma that the eigenvectors $\mathbf{v}_0, \mathbf{v}_1, \ldots, \mathbf{v}_{n-1}$ of $L$ in some sense capture the principal directions of variation of functions on $G$, the boundedness of the eigenvalues $\lambda_0, \lambda_1, \ldots, \lambda_{n-1}$ allow us to construct a whole family of possible kernels of the form

$$K = \sum_{i=0}^{n-1} (r(\lambda_i))^{-1}\,\mathbf{v}_i\,\mathbf{v}_i^{\top}, \tag{61}$$

the only restriction on $r$ being that it must be positive and increasing on $[0, 2]$. A variety of such functions is presented by Smola and Kondor (2003).

The connection between the adjacency matrix of the direct product graph and simultaneous random walks, presented in Section 2, raises the intriguing possibility that the Laplacian might also be useful in defining kernels on graphs, as opposed to just graph vertices. In particular, replacing $W_\times$ in (8) by $\widetilde{\mathsf{L}} \otimes \widetilde{\mathsf{L}}'$ yields an alternate similarity measure between graphs. If $\widetilde{\mathsf{L}} \otimes \widetilde{\mathsf{L}}'$ were the Laplacian of the direct product graph, then this would amount to computing the expectation of the diffusion kernel on the nodes of the product graph under the distribution $\mathrm{vec}(q_\times p_\times^{\top})$.

Unfortunately, the Laplacian of the product graph does not decompose into the Kronecker product of the Laplacian matrices of the constituent graphs, that is, $\widetilde{\mathsf{L}}_\times \neq \widetilde{\mathsf{L}} \otimes \widetilde{\mathsf{L}}'$.

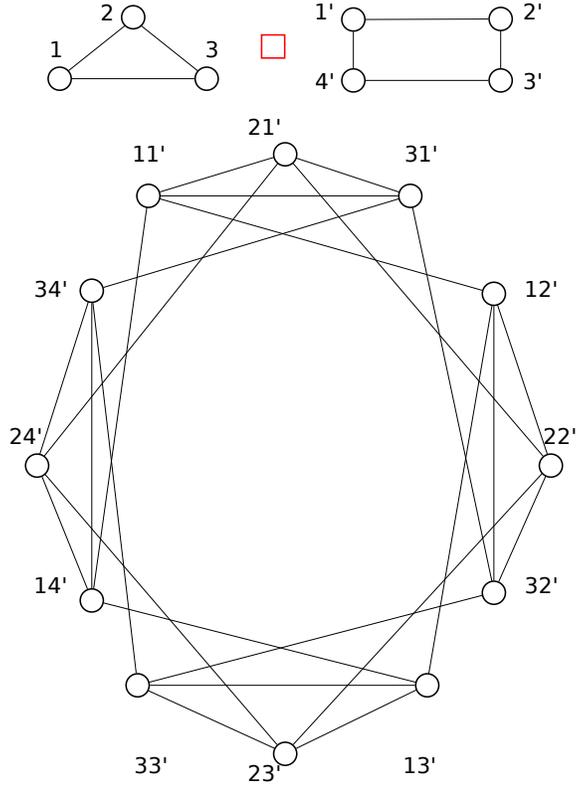

Figure 5: Two graphs (top left & right) and their Cartesian product (bottom). Each node of the Cartesian product graph is labeled with a pair of nodes; an edge exists in the Cartesian product if and only if the corresponding nodes are identical in one and adjacent in the other original graph. For instance, nodes $31'$ and $32'$ are adjacent because they refer to the same node 3 in the first graph, and there is an edge between nodes $1'$ and $2'$ in the second graph.

Therefore, replacing $W_\times$ by $\widetilde{L} \otimes \widetilde{L}'$ leads to a valid *p.s.d.* kernel, but then we lose the physical interpretation relating this to a diffusion process on a product graph. This can be rectified by employing the *Cartesian* product graph instead.

### 7.2 Cartesian Product Graph Kernels

Given two graphs $G(V, E)$ and $G'(V', E')$ (with $|V| = n$ and $|V'| = n'$), the Cartesian product $G_\square$ (Imrich and Klavžar, 2000) is a graph with vertex set

$$V_\square = \{(v_i, v'_{i'}) : v_i \in V, \ v'_{i'} \in V'\}, \tag{62}$$

and edge set

$$E_\square = \{((v_i, v'_{i'}), (v_j, v'_{j'})) : v_i = v_j \text{ and } (v'_{i'}, v'_{j'}) \in E' \lor v'_{i'} = v'_{j'} \text{ and } (v_i, v_j) \in E\} \tag{63}$$

(*cf.* Figure 5). It is easy to verify that $\widetilde{A}_{\square} = \widetilde{A} \oplus \widetilde{A}'$ and $\widetilde{L}_{\square} = \widetilde{L} \oplus \widetilde{L}'$, where $\oplus$ is the Kronecker sum (3). One can now write an analogue of (8) for the Cartesian product graph: Given the weight matrix $W_{\square} := \Phi(L) \oplus \Phi(L')$, initial and stopping probability distributions $p_{\square} := (p \otimes p')$ and $q_{\square} := (q \otimes q')$, and an appropriately chosen discrete measure $\mu$, we can define a kernel on $G$ and $G'$ as

$$k(G, G') := \sum_{k=1}^{\infty} \mu(k) \, q_{\square}^{\top} W_{\square}^{2k} \, p_{\square}. \tag{64}$$

Letting $\Phi(L)^0 = \mathbf{I}$ we show in Appendix B that

**Lemma 9** *If the measure $\mu(k)$ is such that (64) converges, then it defines a valid* p.s.d. *kernel.*

Two things are worth noting here: First, we use $W^{2k}$ instead of $W^k$ as was used in (8). This helps us to overcome the technical difficulty that while $W_{\times}$ is a real matrix $W_{\square}$ is a matrix in RKHS. Although we will not pursue this avenue here, we note in the passing that one can analogously redefine (8) using $W_{\times}^{2k}$. Second, we define $p_{\square}$ (analogously $q_{\square}$) as $p \otimes p'$ instead of the Kronecker sum $\frac{1}{2}(p \oplus p')$, which would also define a valid probability distribution. The problem with the latter formulation is that it would destroy the invariance of the diffusion kernel to permutation of graph nodes, and thus render it far less useful.

## 7.3 Efficient Computation of Cartesian Product Kernels

The techniques for efficiently computing random walk graph kernels via direct product graphs we introduced in Section 3 are equally applicable to the computation of diffusion-based kernels via Cartesian product graphs. In particular, the conjugate gradient (Section 3.2) and fixed-point iteration (Section 3.3) methods can be used without any modification. They will take at most twice as much time on the Cartesian product graph than on the direct product graph, due to the lower sparsity of the Kronecker sum *vs.* the Kronecker product of two sparse matrices.

Our Sylvester equation-based method (Section 3.1) can also be used here: Assume that the weight matrix $W_{\square}$ can be written as

$$W_{\square} = \sum_{l=1}^{d} {}^{l}L \oplus {}^{l}L', \tag{65}$$

where the ${}^{l}L$ and ${}^{l}L'$ are label-filtered normalized Laplacian matrices, which are defined analogously to label-filtered normalized adjacency matrices (*cf.* Section 2.1). The problem of computing the graph kernel (64) with $\mu(k) := \lambda^k$ can be reduced to the problem of solving the Sylvester equation:

$$M = \sum_{i=1}^{d} \lambda \, {}^{i}L' \, M \, \mathbf{I}_A^{\top} + \sum_{i=1}^{d} \lambda \, \mathbf{I}_{A'} \, M \, {}^{i}L^{\top} + M_0, \tag{66}$$

where $\mathrm{vec}(M_0) = p_{\square}$. We begin by *flattening* (66):

$$\mathrm{vec}(M) = \lambda \sum_{i=1}^{d} \mathrm{vec}({}^{i}L' M \, \mathbf{I}_A^{\top}) + \lambda \sum_{i=1}^{d} \mathrm{vec}(\mathbf{I}_{A'} \, M \, {}^{i}L^{\top}) + p_{\square}. \tag{67}$$

Using (77) we can rewrite (67) as

$$(\mathbf{I} - \lambda \sum_{i=1}^{d} {}^{i}L \oplus {}^{i}L') \operatorname{vec}(M) = p_\square, \tag{68}$$

use (65), and solve (68) for $\operatorname{vec}(M)$:

$$\operatorname{vec}(M) = (\mathbf{I} - \lambda W_\square)^{-1} p_\square. \tag{69}$$

Multiplying both sides of (69) by $q_\square^\top$ yields

$$q_\square^\top \operatorname{vec}(M) = q_\square^\top (\mathbf{I} - \lambda W_\square)^{-1} p_\square. \tag{70}$$

The right-hand side of (70) is the Cartesian product kernel (64). Compared to the direct product kernel, the computation will take twice as long because the degree of the generalized Sylvester equation (66) is now $2d$ instead of $d$.

## 7.4 A Deficiency of Diffusion-Based Graph Kernels

Putting everything together, we can construct diffusion-based graph kernels via the Cartesian product graph, and evaluate them efficiently. However, we found the resulting diffusion-based graph kernels to suffer from a troubling deficiency: Recall that $D_{ii} = \sum_j \widetilde{A}_{ij}$, while $\widetilde{L} = D - \widetilde{A}$. This means that $\forall i : [\widetilde{L} \mathbf{e}]_i = D_{ii} - \sum_j \widetilde{A}_{ij} = 0$. In other words, $\mathbf{e}$ is an eigenvector of $\widetilde{L}$ with zero eigenvalue (and consequently $\widetilde{L}$ is rank deficient). Thus for any $k > 0$ we have $\widetilde{L}^k \mathbf{e} = 0$.

In the absence of any prior knowledge about the data on hand, it is, natural to set the initial and stopping probabilities $p_\square$ resp. $q_\square$ to uniform distributions; given graphs $G$ and $G'$ of size $n$ and $n'$, respectively, we set $p_\square = q_\square = \mathbf{e}/(nn')$. The above discussion, however, implies that then $q_\square^\top \widetilde{L}^{2k} p_\square = 0$ for all $k > 0$, and consequently (64) with $W_\square = L_\square$ is uniformly zero.

One might be tempted to create non-uniform $p_\square$ and $q_\square$ based on available properties of $G$ and $G'$, such as the degree of their nodes: $p_\square = q_\square \propto \operatorname{diag}(D) \otimes \operatorname{diag}(D')$. For every such strategy, however, there will be a subclass of graphs (in our example: regular graphs) which yields uniform distributions, and whose members are therefore indistinguishable to diffusion-based kernels. Breaking this uniformity arbitrarily would conversely destroy the permutation invariance of the kernel.

We are forced to conclude that in order to use diffusion-based graph kernels, we must have either a) some prior knowledge about the initial and stopping probabilities on the graph, or b) a rich enough feature representation to ensure that the weight matrix $W_\square$ is not rank deficient. Since none of our datasets satisfy this requirement we do not report experiments on diffusion-based graph kernels.

## 8. Outlook and Discussion

As evidenced by the large number of recent papers, random walk graph kernels and marginalized graph kernels have received considerable research attention. Although the connections between these two kernels were hinted at by Kashima et al. (2004), no effort was made to pursue this further. Our aim in presenting a unified framework to view random walk graph kernels, marginalized kernels on graphs, and geometric kernels on graphs is to highlight the similarities as well as the differences between these approaches. Furthermore, this allows us to use extended linear algebra in an RKHS to efficiently compute these kernels by exploiting structure inherent in these problems.

Although rational kernels have always been viewed as distinct from graph kernels, we showed that in fact these two research areas are closely related. It is our hope that this will facilitate cross-pollination of ideas such as the use of semirings and transducers in defining graph kernels. We also hope that tensor and matrix notation become more prevalent in the transducer community.

It is fair to say that R-convolution kernels are the *mother* of all kernels on structured data. It is enlightening to view various graph kernels as instances of R-convolution kernels since this brings into focus the relevant decomposition used to define a given kernel, and the similarities and differences between various kernels. However, extending R-convolutions to abstract semirings does not always result in a valid *p.s.d.* kernel.

The links between diffusion kernels and generalized random walk kernels on graphs are intriguing. It is fascinating that direct product graphs are linked with random walks, but Cartesian product graphs arise when studying diffusion. Surprisingly, all our efficient computational tricks from the random walk kernels translate to the diffusion-based kernel. As we showed, however, a rank deficiency limits their applicability. Identifying domains where diffusion-based graph kernels are applicable is a subject of future work. It is plausible that $W_\square$ in (64) can be replaced by a spectral function similar to (61). The necessary and sufficient conditions for admissible spectral functions $r(\lambda)$ in this case remain an open question.

As more and more graph-structured data (e.g., molecular structures and protein interaction networks) becomes available in fields such as biology, web data mining, *etc.*, graph classification will gain importance over the coming years. Hence there is a pressing need to speed up the computation of similarity metrics on graphs. We have shown that sparsity, low effective rank, and Kronecker product structure can be exploited to greatly reduce the computational cost of graph kernels; taking advantage of other forms of structure in $W_\times$ remains a computational challenge. Now that the computation of random walk graph kernels is viable for practical problem sizes, it will open the doors for their application in hitherto unexplored domains.

A major deficiency of the random walk graph kernels can be understood by studying (13). The admissible values of the decay parameter $\lambda$ is often dependent on the spectrum of the matrices involved. What this means in practise is that one often resorts to using very low values of $\lambda$. But a small $\lambda$ makes the contributions to the kernel of higher-order terms (corresponding to long walks) negligible. In fact in many applications a naive kernel which simply computes the average kernel between all pairs of edges in the two graphs has performance comparable to the random walk graph kernel.

Trying to rectify this situation by normalizing the matrices involved brings to the fore another phenomenon called *tottering* (Mahé et al., 2004). Roughly speaking tottering implies that short self-repeating walks have a disproportionately large contribution to the kernel value. Consider two adjacent vertices $v$ and $v'$ in a graph. Because of tottering contributions due to walks of the form $v \to v' \to v \to \ldots$ dominate the kernel value. Unfortunately a kernel using self-avoiding walks (walks which do not visit the same vertex twice) cannot be computed in polynomial time.

We do not believe that the last word on graph comparison has been said yet. Thus far, simple decompositions like random walks have been used to compare graphs. This is mainly driven by computational considerations and not by the application domain on hand. The algorithmic challenge of the future is to integrate higher-order structures, such as spanning trees, in graph comparisons, and to compute such kernels efficiently.

## Acknowledgments

We thank Markus Hegland and Tim Sears for enlightening discussions, and Alex Smola for pointing out that the optimal assignment kernel may fail to be *p.s.d.* A short, early version of this work was presented at the NIPS conference (Vishwanathan et al., 2006). The experiments on protein-protein interaction first appeared in Borgwardt et al. (2007).

NICTA is funded by the Australian Government's Backing Australia's Ability and the Centre of Excellence programs. This work is also supported by the IST Program of the European Community, under the FP7 Network of Excellence, ICT-216886-NOE, by the German Ministry for Education, Science, Research and Technology (BMBF) under grant No. 031U112F within the BFAM (Bioinformatics for the Functional Analysis of Mammalian Genomes) project, part of the German Genome Analysis Network (NGFN), and by NIH grant GM063208-05 "Tools and Data Resources in Support of Structural Genomics."

## Appendix A. Extending Linear Algebra to RKHS

It is well known that any symmetric, positive definite kernel $\kappa \colon \mathcal{X} \times \mathcal{X} \to \mathbb{R}$ has a corresponding Hilbert space $\mathcal{H}$ (called the Reproducing Kernel Hilbert Space or RKHS) and a feature map $\phi \colon \mathcal{X} \to \mathcal{H}$ satisfying $\kappa(x, x') = \langle \phi(x), \phi(x') \rangle_{\mathcal{H}}$. The natural extension of this so-called feature map to matrices is $\Phi \colon \mathcal{X}^{n \times m} \to \mathcal{H}^{n \times m}$ defined $[\Phi(A)]_{ij} := \phi(A_{ij})$. In what follows, we use $\Phi$ to lift tensor algebra from $\mathcal{X}$ to $\mathcal{H}$, extending various matrix products to the RKHS, and proving some of their their useful properties. Straightforward extensions via the commutativity properties of the operators have been omitted for the sake of brevity.

### A.1 Matrix Product

**Definition 10** *Let $A \in \mathcal{X}^{n \times m}$, $B \in \mathcal{X}^{m \times p}$, and $C \in \mathbb{R}^{m \times p}$. The matrix products $\Phi(A) \Phi(B) \in \mathbb{R}^{n \times p}$ and $\Phi(A) C \in \mathcal{H}^{n \times p}$ are given by*

$$[\Phi(A)\Phi(B)]_{ik} := \sum_j \langle \phi(A_{ij}), \phi(B_{jk}) \rangle_{\mathcal{H}} \qquad and \qquad [\Phi(A)\, C]_{ik} := \sum_j \phi(A_{ij})\, C_{jk}.$$

It is straightforward to show that the usual properties of matrix multiplication — namely associativity, transpose-commutativity, and distributivity with addition — hold for Definition 10 above, with one exception: associativity does *not* hold if the elements of all three matrices involved belong to the RKHS. In other words, given $A \in \mathcal{X}^{n \times m}$, $B \in \mathcal{X}^{m \times p}$, and $C \in \mathcal{X}^{p \times q}$, in general $[\Phi(A)\Phi(B)]\Phi(C) \neq \Phi(A)[\Phi(B)\Phi(C)]$. The technical difficulty is that

$$\langle \phi(A_{ij}), \phi(B_{jk}) \rangle_{\mathcal{H}} \, \phi(C_{kl}) \neq \phi(A_{ij}) \, \langle \phi(B_{jk}), \phi(C_{kl}) \rangle_{\mathcal{H}} . \tag{71}$$

Further examples of statements like (71), involving properties which not hold when extended to an RKHS, can be found for the other matrix products at (73) and (79) below.

Definition 10 allows us to state a first RKHS extension of the vec(ABC) formula (1):

**Lemma 11** *If $A \in \mathbb{R}^{n \times m}$, $B \in \mathcal{X}^{m \times p}$, and $C \in \mathbb{R}^{p \times q}$, then*

$$\mathrm{vec}(A\, \Phi(B)\, C)) = (C^{\top} \otimes A)\, \mathrm{vec}(\Phi(B)) \ \in \mathcal{X}^{nq \times 1} .$$

**Proof** Analogous to Lemma 13 below. ∎

### A.2 Kronecker Product

**Definition 12** *Let $A \in \mathcal{X}^{n \times m}$ and $B \in \mathcal{X}^{p \times q}$. The Kronecker product $\Phi(A) \otimes \Phi(B) \in \mathbb{R}^{np \times mq}$ is defined as*

$$[\Phi(A) \otimes \Phi(B)]_{(i-1)p+k,(j-1)q+l} := \langle \phi(A_{ij}), \phi(B_{kl}) \rangle_{\mathcal{H}} .$$

Similarly to (71) above, for matrices in an RKHS

$$* \qquad (\Phi(A) \otimes \Phi(B))(\Phi(C) \otimes \Phi(D)) = (\Phi(A)\,\Phi(C)) \otimes (\Phi(B)\,\Phi(D)) \qquad (72)$$

does *not* necessarily hold. The technical problem with (72) is that generally

$$\langle \phi(A_{ir}), \phi(B_{ks})\rangle_{\mathcal{H}} \, \langle \phi(C_{rj}), \phi(D_{sl})\rangle_{\mathcal{H}} \neq \langle \phi(A_{ir}), \phi(C_{rj})\rangle_{\mathcal{H}} \, \langle \phi(B_{ks}), \phi(D_{sl})\rangle_{\mathcal{H}}. \qquad (73)$$

In Section A.3 we show that analogous properties (Lemmas 15 and 16) do hold for the *heterogeneous* Kronecker product between RKHS and real matrices.

Definition 12 gives us a second extension of the vec(ABC) formula (1) to RKHS:

**Lemma 13** *If $A \in \mathcal{X}^{n \times m}$, $B \in \mathbb{R}^{m \times p}$, and $C \in \mathcal{X}^{p \times q}$, then*

$$\mathrm{vec}(\Phi(A)\,B\,\Phi(C)) = (\Phi(C)^{\top} \otimes \Phi(A))\,\mathrm{vec}(B) \ \in \mathbb{R}^{nq \times 1}.$$

**Proof** We begin by rewriting the $k^{\text{th}}$ column of $\Phi(A)B\Phi(C)$ as

$$[\Phi(A)B\Phi(C)]_{*k} = \Phi(A) \sum_{j} B_{*j}\,\phi(C_{jk}) = \sum_{j} \phi(C_{jk})\Phi(A)B_{*j}$$

$$= [\phi(C_{1k})\Phi(A), \phi(C_{2k})\Phi(A), \ldots \phi(C_{nk})\Phi(A)] \underbrace{\begin{bmatrix} B_{*1} \\ B_{*2} \\ \vdots \\ B_{*n} \end{bmatrix}}_{\mathrm{vec}(B)}$$

$$= ([\phi(C_{1k}), \phi(C_{2k}), \ldots \phi(C_{nk})] \otimes \Phi(A))\,\mathrm{vec}(B). \qquad (74)$$

To obtain Lemma 13 we stack up the columns of (74):

$$\mathrm{vec}(\Phi(A)\,B\,\Phi(C)) = \left( \begin{bmatrix} \phi(C_{11}) & \phi(C_{21}) & \ldots & \phi(C_{n1}) \\ \vdots & \vdots & \ddots & \vdots \\ \phi(C_{1n}) & \phi(C_{2n}) & \ldots & \phi(C_{nn}) \end{bmatrix} \otimes \Phi(A) \right) \mathrm{vec}(B)$$

$$= (\Phi(C)^{\top} \otimes \Phi(A))\,\mathrm{vec}(B). \qquad \blacksquare$$

Direct computation of the right-hand side of Lemma 13 requires $nmpq$ kernel evaluations; when $m$, $p$, and $q$ are all $O(n)$ this is $O(n^4)$. If $\mathcal{H}$ is finite-dimensional, however — in other words, if the feature map can be taken to be $\phi \colon \mathcal{X} \to \mathbb{R}^d$ with $d < \infty$ — then the left-hand side of Lemma 13 can be obtained in $O(n^3 d)$ operations. Our efficient computation schemes in Section 3 exploit this observation.

### A.3 Heterogeneous Kronecker Product

**Definition 14** *Let $A \in \mathcal{X}^{n \times m}$ and $B \in \mathbb{R}^{p \times q}$. The heterogeneous Kronecker product $\Phi(A) \otimes B \in \mathcal{X}^{np \times mq}$ is given by*

$$[\Phi(A) \otimes B]_{(i-1)p+k,(j-1)q+l} = \phi(A_{ij}) B_{kl}.$$

Recall that the standard Kronecker product obeys (2); here we prove two extensions:

**Lemma 15** *If $A \in \mathcal{X}^{n \times m}$, $B \in \mathcal{X}^{p \times q}$, $C \in \mathbb{R}^{m \times o}$, and $D \in \mathbb{R}^{q \times r}$, then*

$$(\Phi(A) \otimes \Phi(B))(C \otimes D) = (\Phi(A)\,C) \otimes (\Phi(B)\,D).$$

**Proof** Using the linearity of the inner product we directly verify

$$
\begin{aligned}
[(\Phi(A) \otimes \Phi(B))(C \otimes D)]_{(i-1)p+k,(j-1)q+l} &= \sum_{r,s} \langle \phi(A_{ir}), \phi(B_{ks}) \rangle_{\mathcal{H}}\, C_{rj} D_{sl} \\
&= \left\langle \sum_r \phi(A_{ir}) C_{rj}, \sum_s \phi(B_{ks}) D_{sl} \right\rangle_{\mathcal{H}} \\
&= \langle [\Phi(A)\,C]_{ij}, [\Phi(B)\,D]_{kl} \rangle_{\mathcal{H}} \\
&= [(\Phi(A)\,C) \otimes (\Phi(B)\,D)]_{(i-1)p+k,(j-1)q+l}
\end{aligned}
$$

∎

**Lemma 16** *If $A \in \mathcal{X}^{n \times m}$, $B \in \mathbb{R}^{p \times q}$, $C \in \mathcal{X}^{m \times o}$, and $D \in \mathbb{R}^{q \times r}$, then*

$$(\Phi(A) \otimes B)(\Phi(C) \otimes D) = (\Phi(A)\,\Phi(C)) \otimes (B\,D).$$

**Proof** Using the linearity of the inner product we directly verify

$$
\begin{aligned}
[(\Phi(A) \otimes B)(\Phi(C) \otimes D)]_{(i-1)p+k,(j-1)q+l} &= \sum_{r,s} \langle \phi(A_{ir}) B_{ks}, \phi(C_{rj}) D_{sl} \rangle_{\mathcal{H}} \\
&= \sum_r \langle \phi(A_{ir}), \phi(C_{rj}) \rangle_{\mathcal{H}} \sum_s B_{ks} D_{sl} \\
&= [\Phi(A)\,\Phi(C)]_{ij}\,[B\,D]_{kl} \\
&= [(\Phi(A)\,\Phi(C)) \otimes (B\,D)]_{(i-1)p+k,(j-1)q+l}
\end{aligned}
$$

∎

Using the heterogeneous Kronecker product, we can state four more RKHS extensions of the vec-ABC formula (1):

**Lemma 17** *If $A \in \mathcal{X}^{n \times m}$, $B \in \mathbb{R}^{m \times p}$, and $C \in \mathbb{R}^{p \times q}$, then*

$$\text{vec}(\Phi(A) \, B \, C) = (C^\top \otimes \Phi(A)) \, \text{vec}(B) \ \in \mathcal{X}^{nq \times 1}.$$

**Proof** Analogous to Lemma 13. ■

**Lemma 18** *If $A \in \mathbb{R}^{n \times m}$, $B \in \mathbb{R}^{m \times p}$, and $C \in \mathcal{X}^{p \times q}$, then*

$$\text{vec}(A \, B \, \Phi(C)) = (\Phi(C)^\top \otimes A) \, \text{vec}(B) \ \in \mathcal{X}^{nq \times 1}.$$

**Proof** Analogous to Lemma 13. ■

**Lemma 19** *If $A \in \mathcal{X}^{n \times m}$, $B \in \mathcal{X}^{m \times p}$, and $C \in \mathbb{R}^{p \times q}$, then*

$$\text{vec}(\Phi(A) \, \Phi(B) \, C) = (C^\top \otimes \Phi(A)) \, \text{vec}(\Phi(B)) \ \in \mathbb{R}^{nq \times 1}.$$

**Proof** Analogous to Lemma 13. ■

**Lemma 20** *If $A \in \mathbb{R}^{n \times m}$, $B \in \mathcal{X}^{m \times p}$, and $C \in \mathcal{X}^{p \times q}$, then*

$$\text{vec}(A \, \Phi(B) \, \Phi(C)) = (\Phi(C)^\top \otimes A) \, \text{vec}(\Phi(B)) \ \in \mathbb{R}^{nq \times 1}.$$

**Proof** Analogous to Lemma 13. ■

### A.4 Kronecker Sum

Unlike the Kronecker product, the Kronecker sum of two matrices in an RKHS is also an matrix in the RKHS. From Definition 1 and (3) we find that

$$[A \oplus B]_{(i-1)p+k, (j-1)q+l} := A_{ij}\delta_{kl} + \delta_{ij}B_{kl}. \tag{75}$$

We can extend (75) to RKHS, defining analogously:

**Definition 21** *Let $A \in \mathcal{X}^{n \times m}$ and $B \in \mathcal{X}^{p \times q}$. The Kronecker sum $\Phi(A) \oplus \Phi(B) \in \mathcal{X}^{np \times mq}$ is defined as*

$$[\Phi(A) \oplus \Phi(B)]_{(i-1)p+k, (j-1)q+l} := \phi(A_{ij})\delta_{kl} + \delta_{ij}\phi(B_{kl}).$$

In other words, in an RKHS the Kronecker sum is defined just as in (3):

$$\Phi(A) \oplus \Phi(B) = \Phi(A) \otimes \mathbf{I}_B + \mathbf{I}_A \otimes \Phi(B), \tag{76}$$

where $\mathbf{I}_M$ denotes the real-valued identity matrix of the same dimensions (not necessarily square) as matrix $M$. In accordance with Definition 14, the result of (76) is an RKHS matrix.

The equivalent of the vec-ABC formula (1) for Kronecker sums is:

$$
\begin{aligned}
(A \oplus B)\operatorname{vec}(C) &= (A \otimes \mathbf{I}_B + \mathbf{I}_A \otimes B)\operatorname{vec}(C) \\
&= (A \otimes \mathbf{I}_B)\operatorname{vec}(C) + (\mathbf{I}_A \otimes B)\operatorname{vec}(C) \\
&= \operatorname{vec}(\mathbf{I}_B\, C\, A^\top) + \operatorname{vec}(BC\, \mathbf{I}_A^\top) \\
&= \operatorname{vec}(\mathbf{I}_B\, C\, A^\top + BC\, \mathbf{I}_A^\top).
\end{aligned}
\tag{77}
$$

This also works for matrices in an RKHS:

**Lemma 22** *If $A \in \mathcal{X}^{n \times m}$, $B \in \mathcal{X}^{p \times q}$, and $C \in \mathcal{X}^{q \times m}$, then*

$$(\Phi(A) \oplus \Phi(B))\operatorname{vec}(\Phi(C)) = \operatorname{vec}(\mathbf{I}_B\, \Phi(C)\, \Phi(A)^\top + \Phi(B)\, \Phi(C)\, \mathbf{I}_A^\top) \ \in \mathbb{R}^{np \times 1}.$$

**Proof**  Analogous to (77), using Lemmas 19 and 20. ∎

Furthermore, we have two valid heterogeneous forms that map into the RKHS:

**Lemma 23** *If $A \in \mathcal{X}^{n \times m}$, $B \in \mathcal{X}^{p \times q}$, and $C \in \mathbb{R}^{q \times m}$, then*

$$(\Phi(A) \oplus \Phi(B))\operatorname{vec}(C) = \operatorname{vec}(\mathbf{I}_B\, C\, \Phi(A)^\top + \Phi(B)\, C\, \mathbf{I}_A^\top) \ \in \mathcal{X}^{np \times 1}.$$

**Proof**  Analogous to (77), using Lemmas 17 and 18. ∎

**Lemma 24** *If $A \in \mathbb{R}^{n \times m}$, $B \in \mathbb{R}^{p \times q}$, and $C \in \mathcal{X}^{q \times m}$, then*

$$(A \oplus B)\operatorname{vec}(\Phi(C)) = \operatorname{vec}(\mathbf{I}_B\, \Phi(C)\, A^\top + B\, \Phi(C)\, \mathbf{I}_A^\top) \ \in \mathcal{X}^{np \times 1}.$$

**Proof**  Analogous to (77), using Lemma 11. ∎

### A.5 Hadamard Product

While the extension of the Hadamard (element-wise) product to an RKHS is not required to implement our fast graph kernels, the reader may find it interesting in its own right.

**Definition 25** *Let $A, B \in \mathcal{X}^{n \times m}$ and $C \in \mathbb{R}^{n \times m}$. The Hadamard products $\Phi(A) \odot \Phi(B) \in \mathbb{R}^{n \times m}$ and $\Phi(A) \odot C \in \mathcal{H}^{n \times m}$ are given by*

$$[\Phi(A) \odot \Phi(B)]_{ij} = \langle \phi(A_{ij}), \phi(B_{ij}) \rangle_{\mathcal{H}} \quad and \quad [\Phi(A) \odot C]_{ij} = \phi(A_{ij}) \, C_{ij}.$$

We prove two extensions of (4):

**Lemma 26** *If $A \in \mathcal{X}^{n \times m}$, $B \in \mathcal{X}^{p \times q}$, $C \in \mathbb{R}^{n \times m}$, and $D \in \mathbb{R}^{p \times q}$, then*

$$(\Phi(A) \otimes \Phi(B)) \odot (C \otimes D) = (\Phi(A) \odot C) \otimes (\Phi(B) \odot D).$$

**Proof**  Using the linearity of the inner product we directly verify

$$\begin{aligned}
[(\Phi(A) \otimes \Phi(B)) \odot (C \otimes D)]_{(i-1)p+k,(j-1)q+l} &= \langle \phi(A_{ij}), \phi(B_{kl}) \rangle_{\mathcal{H}} \, C_{ij} D_{kl} \\
&= \langle \phi(A_{ij}) C_{ij}, \phi(B_{kl}) D_{kl} \rangle_{\mathcal{H}} \\
&= \langle [\Phi(A) \odot C]_{ij}, [\Phi(B) \odot D]_{kl} \rangle_{\mathcal{H}} \\
&= [(\Phi(A) \odot C) \otimes (\Phi(B) \odot D)]_{(i-1)p+k,(j-1)q+l}
\end{aligned}$$

∎

**Lemma 27** *If $A \in \mathcal{X}^{n \times m}$, $B \in \mathbb{R}^{p \times q}$, $C \in \mathcal{X}^{n \times m}$, and $D \in \mathbb{R}^{p \times q}$, then*

$$(\Phi(A) \otimes B) \odot (\Phi(C) \otimes D) = (\Phi(A) \odot \Phi(C)) \otimes (B \odot D).$$

**Proof**  Using the linearity of the inner product we directly verify

$$\begin{aligned}
[(\Phi(A) \otimes B) \odot (\Phi(C) \otimes D)]_{(i-1)p+k,(j-1)q+l} &= \langle \phi(A_{ij}) B_{kl}, \phi(C_{ij}) D_{kl} \rangle_{\mathcal{H}} \\
&= \langle \phi(A_{ij}), \phi(C_{ij}) \rangle_{\mathcal{H}} \, B_{kl} D_{kl} \\
&= [\Phi(A) \odot \Phi(C)]_{ij} [B \odot D]_{kl} \\
&= [(\Phi(A) \odot \Phi(C)) \otimes (B \odot D)]_{(i-1)p+k,(j-1)q+l}
\end{aligned}$$

∎

As before,

$$* \qquad (\Phi(A) \otimes \Phi(B)) \odot (\Phi(C) \otimes \Phi(D)) = (\Phi(A) \odot \Phi(C)) \otimes (\Phi(B) \odot \Phi(D)) \qquad (78)$$

does *not* necessarily hold, the difficulty with (78) being that in general,

$$\langle \phi(A_{ij}), \phi(B_{kl}) \rangle_{\mathcal{H}} \langle \phi(C_{ij}), \phi(D_{kl}) \rangle_{\mathcal{H}} \neq \langle \phi(A_{ij}), \phi(C_{ij}) \rangle_{\mathcal{H}} \langle \phi(B_{kl}), \phi(D_{kl}) \rangle_{\mathcal{H}}. \qquad (79)$$

## Appendix B. Cartesian Product Kernels: Proof of Lemma 9

We first prove the following technical lemma:

**Lemma 28** $\quad \forall \ k \in \mathbb{N}: \quad W_\square^k p_\square = \sum_{i=0}^k \binom{k}{i} \mathrm{vec}[\Phi(L')^{k-i} p' (\Phi(L)^i p)^\top].$

**Proof** By induction over $k$. Base case: $k = 1$. Recall that $\binom{1}{0} = \binom{1}{1} = 1$. Using $p_\square = p \otimes p' = \mathrm{vec}(p'p^\top)$, the definition of $W_\square$, and Lemma 23, we have

$$W_\square \, p_\square = (\Phi(L) \oplus \Phi(L')) \, \mathrm{vec}(p'p^\top)$$
$$= \mathrm{vec}(p'p^\top \Phi(L)^\top + \Phi(L')p'p^\top)$$
$$= \sum_{i=0}^1 \binom{1}{i} \mathrm{vec}[\Phi(L')^{k-i} p' (\Phi(L)^i p)^\top]$$

Induction from $k$ to $k+1$: Using the induction assumption and Lemma 23, we have

$$W_\square^{k+1} p_\square \ = \ (\Phi(L) \oplus \Phi(L')) \sum_{i=0}^k \binom{k}{i} \mathrm{vec}[\Phi(L')^{k-i} p' (\Phi(L)^i p)^\top]$$

$$= \ \sum_{i=0}^k \binom{k}{i} \mathrm{vec}[\Phi(L')^{k-i} p' (\Phi(L)^{i+1} p)^\top \ + \ \Phi(L')^{k-i+1} p' (\Phi(L)^i p)^\top]$$

$$= \ \sum_{j=1}^{k+1} \binom{k}{j-1} \mathrm{vec}[\Phi(L')^{k-j+1} p' (\Phi(L)^j p)^\top] \tag{80}$$

$$+ \ \sum_{i=0}^k \binom{k}{i} \mathrm{vec}[\Phi(L')^{k-i+1} p' (\Phi(L)^i p)^\top],$$

where $j := i + 1$. Pulling out the terms for $j = k+1$ and $i = 0$, we can write (80) as

$$W_\square^{k+1} p_\square \ = \ \mathrm{vec}[p' (\Phi(L)^{k+1} p)^\top] + \mathrm{vec}[\Phi(L')^{k+1} p' \, p^\top]$$

$$+ \sum_{i=1}^k \left[ \binom{k}{i-1} + \binom{k}{i} \right] \mathrm{vec}[\Phi(L')^{k-i+1} p' (\Phi(L)^i p)^\top]. \tag{81}$$

Using the well-known identity $\binom{k+1}{i} = \binom{k}{i-1} + \binom{k}{i}$ (*e.g.*, Abramowitz and Stegun, 1965, Section 24.1.1), and $\binom{k+1}{0} = \binom{k+1}{k+1} = 1$, we can finally rewrite (81) to yield

$$W_\square^{k+1} p_\square \ = \ \binom{k+1}{0} \mathrm{vec}[\Phi(L')^{k+1} p' \, p^\top] \ + \ \binom{k+1}{k+1} \mathrm{vec}[p' (\Phi(L)^{k+1} p)^\top]$$

$$+ \sum_{i=1}^k \binom{k+1}{i} \mathrm{vec}[\Phi(L')^{k-i+1} p' (\Phi(L)^i p)^\top]$$

$$= \ \sum_{i=0}^{k+1} \binom{k+1}{i} \mathrm{vec}[\Phi(L')^{k+1-i} p' (\Phi(L)^i p)^\top], \tag{82}$$

which has the form required by the induction.  ■

We are now positioned to prove Lemma 9:

**Proof**  Using Lemmas 23 and 28, we have

$$q_\square^\top W_\square^{2k} p_\square = (q \otimes q') \sum_{i=0}^{2k} \binom{2k}{i} \text{vec}[\Phi(L')^{2k-i} p' (\Phi(L)^i p)^\top]$$

$$= \sum_{i=0}^{2k} \binom{2k}{i} \text{vec}[q'^\top \Phi(L')^{2k-i} p' (\Phi(L)^i p)^\top q]$$

$$= \sum_{i=0}^{2k} \binom{2k}{i} \underbrace{(q^\top \Phi(L)^i p)^\top}_{\rho(G)^\top} \underbrace{(q'^\top \Phi(L')^{2k-i} p')}_{\rho(G')}. \tag{83}$$

Each individual term of (83) equals $\rho(G)^\top \rho(G')$ for some function $\rho$, and is therefore a valid *p.s.d.* kernel. Because the class of *p.s.d.* kernels is closed under non-negative linear combinations (Berg et al., 1984), the above sum of kernels with positive coefficients $\binom{2k}{i}$ is a valid *p.s.d.* kernel.  ■